\documentclass[10pt,twocolumn,letterpaper]{article}
\usepackage{subfiles}

\usepackage[subpreambles=true]{standalone}
\usepackage{cvpr}              

\usepackage{graphicx}
\usepackage{booktabs}

\usepackage{cite}
\usepackage{amsmath,amssymb,amsfonts}

\usepackage{algorithmic}
\usepackage{textcomp}
\usepackage{epsfig}
\usepackage{amsmath,amssymb,mathtools, xfrac}
\usepackage{rotating,multirow,array}
\usepackage{comment}
\usepackage{hhline}
\usepackage[table]{xcolor}
\usepackage{url}
\usepackage[ruled,vlined]{algorithm2e}
\usepackage{lipsum}
\usepackage{bm}
\usepackage{setspace}
\usepackage{makecell}
\usepackage{cellspace}
\usepackage{pifont}
\usepackage{float}
\usepackage{makecell}

\usepackage{titling}

\usepackage{array}
\newcolumntype{P}[1]{>{\centering\arraybackslash}p{#1}}
\newcolumntype{M}[1]{>{\centering\arraybackslash}m{#1}}

\usepackage[section]{placeins}

\definecolor{red}{rgb}{0.7,0,0}
\definecolor{green}{rgb}{0.0,0.5,0}
\definecolor{blue}{rgb}{0.00,0.00,0.75}
\definecolor{orange}{rgb}{0.72,0.22,0.06}
\definecolor{purple}{rgb}{0.6,0.0,0.6}
\definecolor{pink}{rgb}{1,0.03,0.5}

\usepackage{lipsum}
\newcommand\blfootnote[1]{
\begingroup
\renewcommand\thefootnote{}\footnote{#1}
\addtocounter{footnote}{-1}
\endgroup
}

\def\@fnsymbol#1{\ensuremath{\ifcase#1\or *\or \dagger\or \ddagger\or
   \mathsection\or \mathparagraph\or \|\or **\or \dagger\dagger
   \or \ddagger\ddagger \else\@ctrerr\fi}}
\newcommand{\ssymbol}[1]{^{\@fnsymbol{#1}}}

\newcommand*{\affaddr}[1]{#1}

\usepackage[pagebackref,breaklinks,colorlinks]{hyperref}

\usepackage[capitalize]{cleveref}
\crefname{section}{Sec.}{Secs.}
\Crefname{section}{Section}{Sections}
\Crefname{table}{Table}{Tables}
\crefname{table}{Tab.}{Tabs.}

\begin{document}

\title{Revisiting Image Deblurring with an Efficient ConvNet}
\date{}
\author{
Lingyan Ruan \space\space\space\space   Mojtaba Bemana \space\space\space\space      Hans-peter Seidel\space\space\space\space     Karol Myszkowski \space\space Bin Chen$\ssymbol{1}$ \\
\affaddr{\space Max-Planck-Institut für Informatik}\space\space\space\space\\
\affaddr{\href{https://github.com/lingyanruan/LaKDNet}{\normalsize{https://github.com/lingyanruan/LaKDNet}}}
}

\maketitle

\blfootnote{$*$  denotes corresponding author.}

\begin{abstract}
Image deblurring aims to recover the latent sharp image from its blurry counterpart and has a wide range of applications in computer vision.
The Convolution Neural Networks (CNNs) have performed well in this domain for many years, and until recently an alternative network architecture, namely Transformer, has demonstrated even stronger performance. One can attribute its superiority to the multi-head self-attention (MHSA) mechanism, which offers a larger receptive field and better input content adaptability than CNNs. However, as MHSA demands high computational costs that grow quadratically with respect to the input resolution, it becomes impractical for high-resolution image deblurring tasks. In this work, we propose a unified lightweight CNN network that features a large effective receptive field (ERF) and demonstrates comparable or even better performance than Transformers while bearing less computational costs. Our key design is an efficient CNN block dubbed LaKD, equipped with a large kernel depth-wise convolution and spatial-channel mixing structure, attaining comparable or larger ERF than Transformers but with a smaller parameter scale. Specifically, we achieve +0.17dB / +0.43dB PSNR over the state-of-the-art Restormer on defocus / motion deblurring benchmark datasets with 32\% fewer parameters and 39\% fewer MACs. Extensive experiments demonstrate the superior performance of our network and the effectiveness of each module.  
Furthermore, we propose a compact and intuitive ERFMeter metric that quantitatively characterizes ERF, and shows a high correlation to the network performance. We hope this work can inspire the research community to further explore the pros and cons of CNN and Transformer architectures beyond image deblurring tasks.

\end{abstract}


\vspace{0.1cm}
\section{Introduction}
\label{sec:intro}

Image deblurring plays a major role in the low-level vision realm, especially in the digital age where the camera, as one of the essential parts, has been integrated into almost all types of personal electronic devices. Recovering the latent sharp image from its blurred counterpart has immediate applications on consumer-level electronics and potential benefits to a wide range of vision tasks like object detection \cite{Redmon_2016_CVPR}, image classification \cite{li2022survey}, text recognition \cite{lyu2018mask}, as well as surveillance \cite{thorpe2013coprime} and autonomous driving systems \cite{franke2000real}.
Traditional algorithms depend on blur kernel estimation and blind deconvolution with priors or regularizers to restore sharp images from the observed blurry version~\cite{sandler2018mobilenetv2}. Even though significant progress has been made, the deblurring performance is still limited and tends to introduce unwanted artifacts~\cite{yuan2007image}.
In the last two decades, CNNs have become a promising tool for image deblurring tasks. Given a large dataset, CNNs have the ability to learn the corresponding priors, which then can be used for image deblurring at inference \cite{nah2017deep,zhang2019deep,zamir2021multi,kim2022mssnet,cho2021rethinking,abuolaim2020defocus,lee2021iterative,son2021single,ruan2022learning}, showing high efficiency and generalization ability~\cite{sermanet2013overfeat}.
While on the one hand, the inherent inductive biases contribute to the efficiency of the network; on the other hand, it limits the network's ability to model long-range spatial dependencies.

\begin{figure}[t]
    \setlength{\belowcaptionskip}{0.3cm}
	\centering
	\includegraphics[width=\linewidth]{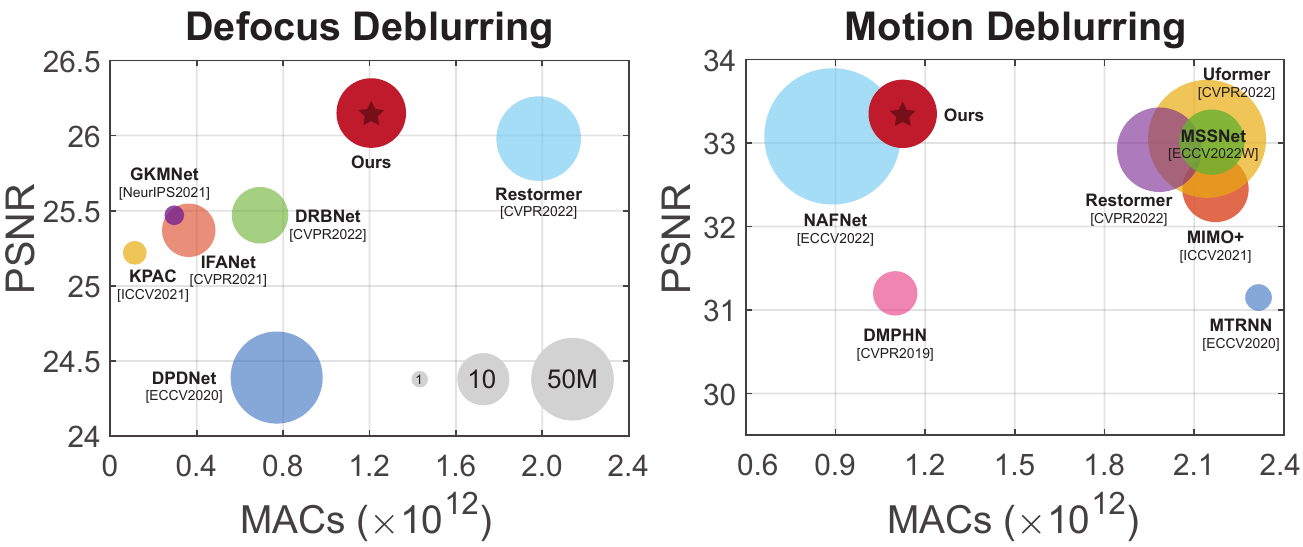}
	\caption{Motion and defocus deblurring results (PSNR) \vs parameters (M, disk size) \vs computational cost (MACs).  Our method achieves the SOTA performance while maintaining efficiency.}
	\label{fig:drb}
\end{figure}

Recently, Transformer, an alternative structure, appears to alleviate the constraints of CNN, and shows compelling performance on several tasks in natural language processing \cite{bahdanau2014neural, vaswani2017attention, devlin2018bert, brown2020language} and computer vision \cite{touvron2021training, dosovitskiy2020image,yuan2022volo,carion2020end,zhu2021deformable}. While the MHSA module in the Transformer solves the long-range spatial dependencies problem, it increases computational complexity. This situation worsens on high-resolution images, where the computational complexity grows quadratically. Although there have been some attempts to relieve the computational burden by bringing back the inductive bias of CNNs ~\cite{vaswani2021scaling, chu2021twins, liu2021swin}, the computational expense is still considerable.
Until recently, Restormer \cite{zamir2022restormer}, in contrast to \cite{liu2021swin,wang2022uformer}, applied self-attention (SA) across the feature dimension instead of the spatial dimension, reducing the computational loads to some extent. Nevertheless, all the Transformer-based approaches benefit from the MHSA mechanism, which is believed to be mainly responsible for creating the large receptive field \cite{raghu2021vision}. A question naturally raises: is it possible to design a pure CNN module capable of approaching a large receptive field with comparable performance to Transformers? Recent works \cite{liu2022convnet,tolstikhin2021mlp,trockman2022patches} resort to the existing network structures, \eg ResNet \cite{he2016deep}, or MobileNet V2 \cite{sandler2018mobilenetv2}, with several modifications, such as group convolution~\cite{chollet2017xception}, inverted bottleneck~\cite{sandler2018mobilenetv2}, or large kernel \cite{ding2022scaling,liu2022more}, demonstrating competitive performance on par with Transformers on a similar model scale. Particularly, RepLKNet \cite{ding2022scaling} and SLaK \cite{liu2022more} build pure CNN models with a focus on increasing ERF using kernel sizes as large as $31 \times 31$ and $51 \times 51$, respectively. While they achieve comparable performance to the Transformer, such explorations of large kernel CNNs are limited to the image classification task. Unlike image classification, which tends to address images with relatively low resolution, image deblurring usually deals with high-resolution inputs, thus imposing further challenges on network architecture design. Whether increasing ERF or equipping CNNs with a large kernel size impacts image deblurring quality has yet to be determined. In this paper, we explore in depth the effect of ERF and large kernel convolution on image deblurring and devise a pure CNN architecture with a block called LaKD, consisting of a large kernel depth-wise convolution and a spatial-channel mixing mechanism.
Moreover, to quantify ERF's influence, we suggest an ERF evaluation method dubbed ERFMeter, presenting a close correlation to the network performance with Pearson correlation coefficient $r=0.8$ when evaluating a large number of networks. Overall, our contributions can be summarized as follows:
\begin{itemize}
 \setlength\itemsep{0.01cm}

  \item To the best of our knowledge, we are the first to investigate the ERF of existing U-Net shape works for motion and defocus debluring and associate their network performance with the ERF. We then present a quantified metric of the ERF, suggesting a more intuitive way to interpret their relationship across different structures. 
  \item We propose a pure CNN structure being able to reach a large ERF, outperforming transformer strucutre (Restormer / Uformer) on high-resolution image deblurring at a small computational cost, specifically sparing up to 66.4\% / 34.6\% parameters and 47.5\% / 43.3\% MACs. 
  \item Extensive experimental and ablation results demonstrate the effectiveness of our method on various benchmark datasets in terms of motion and defocus deblurring. 
   
\end{itemize}

\section{Related Work}
\label{sec:related_work}

\paragraph{Image Deblurring} \; Image deblurring is a long-standing research issue aiming to recover the latent fine textures from its observed version being corrupted by motion \cite{nah2017deep} or defocus \cite{abuolaim2020defocus} blur. Conventional blind image deblurring algorithms often start with kernel estimation, followed by non-blind deconvolution algorithms \cite{karaali2017edge,cho2009fast,shan2008high,xu2010two}. However, these two-step strategies are less effective in terms of quality and computational cost due to the error propagation occurring in their iterative optimization procedure.  Later, CNN-based methods predominate in the image deblurring task by offering end-to-end solutions \cite{lee2021iterative,son2021single,abuolaim2020defocus,ruan2022learning,quan2021gaussian,nah2017deep,zhang2019deep,zamir2021multi,cho2021rethinking,kim2022mssnet}, and efficiently achieving remarkable results, most of which are tailored to a specific type of blur - motion or defocus. Transformer-based structures (\eg Uformer \cite{wang2022uformer}, Restormer \cite{zamir2022restormer}) demonstrate strong ability on various image restoration tasks, including motion and defocus deblurring. However, these unified structures require multi-head self-attention \cite{vaswani2017attention}, resulting in a heavy computational load despite some efforts (\eg narrow down the attention window \cite{wang2022uformer,liu2021swin} or divide the image into patches \cite{chen2021pre} for MHSA) to reduce the costs. In this paper, we explore a pure CNN structure, in contrast to SA, to attain competitive performance on par with or better than the Transformer on a much smaller computation budget.

\begin{figure*}[th]
    \setlength{\abovecaptionskip}{0.1cm}
    \setlength{\belowcaptionskip}{0.1cm}
	\centering
	\includegraphics[width=\linewidth]{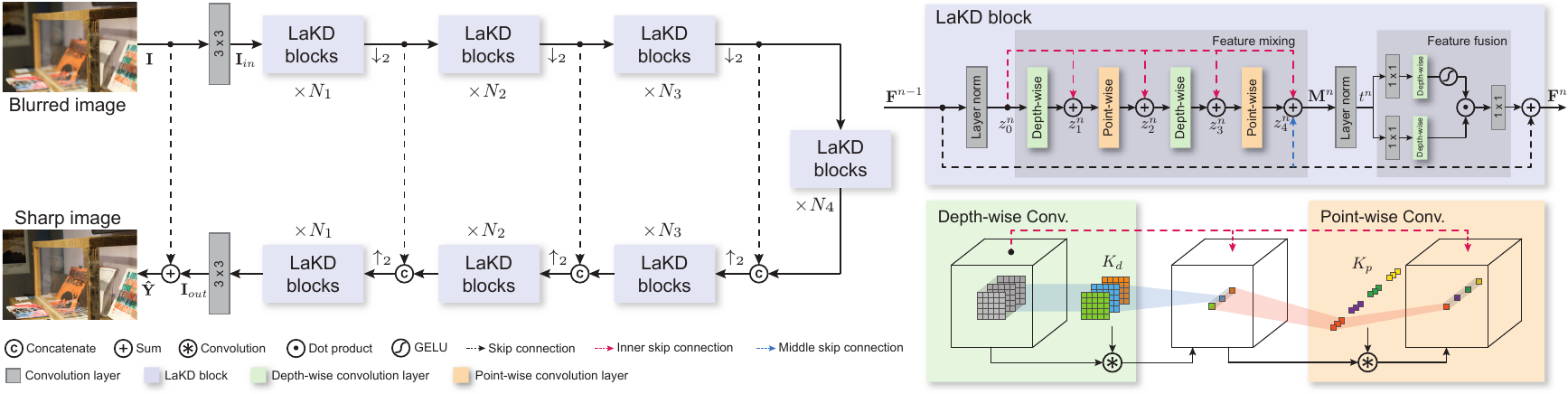}
	\caption{Our U-shape network architecture consists of 4-level symmetric encoder-decoder modules, with each level composed of $N$ LaKD blocks. The LaKD block (top right) contains a feature mixing module and a feature fusion module, where the feature mixing module has depth-wise and point-wise convolution repeated twice, whereas the feature fusion module has only a depth-wise convolution layer with $3\times3$ kernel in a gating mechanism \cite{dauphin2017language}. The details are better viewed when zooming in.}
	\label{fig:network}
    \vspace{-0.5cm}
\end{figure*}

\vspace{0.08cm}
\noindent \textbf{Transformer} \;
The Transformer architecture first took the natural language processing (NLP) community ~\cite{bahdanau2014neural, vaswani2017attention, devlin2018bert, brown2020language} by storm. Later, Dosovitskiy et al. introduced \emph{Vision Transformers (ViTs)}, allowing applications of Transformers on image data~\cite{dosovitskiy2020image}. Since then, it has gained a dominant position in a very short time on a broad range of computer vision tasks \cite{touvron2021training, dosovitskiy2020image,yuan2022volo,carion2020end,xie2021segformer, zheng2021rethinking,liang2021swinir,yang2020learning,wang2022uformer, zamir2022restormer,zhu2021deformable}.
Unlike the conventional CNN that has an inherent inductive bias to the local receptive field, Transformer architectures with a self-attention mechanism allow each pixel to interact with all other pixels in a given patch, yielding a global receptive field and long-range feature dependencies \cite{vaswani2017attention}. Moreover, the convolution filters in self-attention modules are estimated on-the-fly according to the input content, outperforming conventional CNNs, especially at the inference stage~\cite{khan2022transformers}. However, the intensive computational cost restricts applications of the Transformer on high-resolution image data. Thus giving way to the rising trend of new architectures that borrow useful properties from CNNs, and at the same time, avoid their shortcomings~\cite{wang2021pyramid, yuan2021tokens, liu2021swin}. Swin Transformer~\cite{liu2021swin} re-introduces the inductive bias of CNNs to Transformers using a sliding window in a hierarchical architecture to reduce the parameter count and computational cost. This structure has been successfully used as a general-purpose backbone on multiple computer vision tasks~\cite{liu2021swin, liang2021swinir, strudel2021segmenter}. 

\noindent \textbf{Large Kernels} \; Following a similar spirit, some works explore the inverse approach, introducing the merits of Transformer to CNN. Attributing the superiority of ViTs to their large receptive field, researchers resorted back to pure CNNs equipped with larger kernels. This is a well-known older technique; some early works have used large kernel convolution neural networks \cite{krizhevsky2017imagenet} for image classification, such as $11 \times 11$ or $5\times5$ , but since the advent of VGG \cite{simonyan2014very}, they were overtaken by multiple stacked small kernels ($3 \times 3$), which have fewer parameters and enable efficient training. Recent works, influenced by the global attention property of Transformers and MLPs, re-investigate the large convolutional kernels. ConvNeXt \cite{liu2022convnet} adapts the existing ResNet architecture with $7 \times 7$ kernels, RepLKNet \cite{ding2022scaling} and SLaKNet \cite{liu2022more} scale up their kernels to $31 \times 31$ and $51 \times 51$, both of which are purely convolutional neural networks, while achieving performance on par with or even better than Transformers \cite{liu2021swin} on image classification and a few downstream vision tasks. However, little effort has been made to explore the impact of large kernels in image restoration tasks. We investigate the ERF of existing works and propose a unified full CNN structure with large kernels aiming for a large ERF that achieves competitive performance while maintaining efficiency with much fewer parameters and lower computational cost.  


\section{Methodology}
\label{sec:method}
Our method aims to develop an efficient pure CNN model to restore sharp, high-resolution image $\mathbf{\hat{Y}}$ from their blurry version $\mathbf{I}$. To maintain efficiency while attaining a large ERF, we resort to depth-wise convolution with uncommonly large kernels, in contrast to self-attention, to model long-range pixel dependencies. In this section, we introduce the overall structure dubbed $\mathsf{LaKDNet}$ and then provide the details of the proposed basic LaKD block.

\subsection{Overall architecture}
\label{sec:method:network_architecture}
The overall architecture is a U-shape hierarchical network \cite{ronneberger2015u}. It consists of 4-level symmetric encoder-decoder modules, with each level composed of $N$ LaKD blocks, where $N\in\{N_1, N_2, N_3, N_4\}$. Given an input image $\mathbf{I} \in{\mathbb{R}}^{H \times W \times 3}$ with height ($H$) and width ($W$), our network first extracts low-level features $\mathbf{I}_{in} \in{\mathbb{R}}^{H \times W \times C}$ with $C$ channels using a convolutional layer. Then, it is fed into our encoder-decoder structure for blur removal, followed by another convolution layer to recover the features $\mathbf{I}_{out} \in{\mathbb{R}}^{H \times W \times C}$. We apply pixel unshuffle/shuffle for downsampling and upsampling respectively~\cite{shi2016real, zamir2022restormer}. Finally, we skip-connect $\mathbf{I}$ to $\mathbf{I}_{out}$ and produce the sharp image $\mathbf{\hat{Y}}$, forming a global residual structure that is expressed as $\mathbf{\hat{Y}} = \mathbf{I} + \mathsf{LaKDNet}(\mathbf{I})$.

\subsection{LaKD Block}
\label{sec:method:lakd_block}

The motivation behind the LaKD block design is to explore the local and global dependency, as well as a large ERF, in a fully convolutional manner. 
It has two submodules -- feature mixer and feature fusion indicated in Fig. \ref{fig:network}. 
The feature mixer module is similar in spirit to depth-wise separable convolution, but at initial stages employs unusually large kernel sizes (\eg $9 \times 9$) followed by point-wise convolution with a kernel size of $1 \times 1$, along with the inner shortcut between them. 
Unlike the standard convolution layers, which mix spatial and channel dimensions simultaneously through 3D filters, our feature mixer acts separately on spatial intra-channel and depth-wise inter-channel features.
This allows distant spatial location mixture, which combined with large kernel sizes, leads to large ERF (Sec.~\ref{sec:experiment:ablation_study}).
Our design is inspired by MLP-mixer \cite{tolstikhin2021mlp} and ConvMixer \cite{trockman2022patches} which also separately mix spatial and channel dimensions.  
The former advocates the significance of multi-layer perceptrons (MLPs) and the latter focuses on the effectiveness of patch embedding for vision tasks, while we aim for correlating the feature mixer module to the ERF and the restoration performance. Our feature fusion module consists of depth-wise convolution layers with $3 \times 3$  kernels for efficient local information encoding. Similar to \cite{zamir2022restormer}, we employ the gating mechanism \cite{dauphin2017language} that specifically adds an extra path followed by GELU activation function \cite{hendrycks2016gaussian} as the gate, in order to effectively propagate and fuse features. We first introduce our feature mixing module that outputs the feature $\mathbf{M}^n$ in the $n$th LaKD block as follows:
\begin{equation}
    \mathbf{M}^n = \mathbf{F}^{n-1} + z^n_4,
\end{equation}
where $\mathbf{F}^{n-1}$ is the output of the feature fusion module in the $n-1$th LaKD block and $1<n\le N$. The intermediate feature $z_k$ is recursively calculated as:
\begin{equation}
z^n_{k+1} = z^n_{0} + g(z^n_{k}), \quad
g = 
\begin{cases}
\mathsf{depthwise}, & \text{if}\ k =1, 3 \\
\mathsf{pointwise}, & \text{if}\ k =2, 4
\end{cases}\\
\label{eq:feat_mix}
\end{equation}
where the $\mathsf{depthwise}$ and  $\mathsf{pointwise}$ notations represent the depth-wise and point-wise convolutions, respectively and $z^n_0 = \mathsf{LN}(\mathbf{F}^{n-1})$, where $\mathsf{LN}$ is a layer normalization as shown in Fig.~\ref{fig:network}.

Next, we formulate the output feature $\mathbf{F}^n$ from our feature fusion process as:
\begin{equation}
    \mathbf{F}^n = \mathbf{F}^{n-1} + \mathsf{LN}\{\alpha[g(W_1(t^n))]\odot g(W_2(t^n))\},
\end{equation}
where $t^n = \mathsf{LN}(\mathbf{M}^n)$, $W_1$ and $W_2$ are two separate $1\times1$ convolution layers which are then combined as shown in Fig.~\ref{fig:network} using an element-wise multiplication denoted as $\odot$ and followed by a GELU activation $\alpha$. Here, $g$ only applies depth-wise convolution with $3\times3$ kernel.
Our network includes LaKD blocks distributed in a hierarchical manner, allowing a large ERF, and contributing significantly to the fine details restoration. Note that our design shares a similar U-Net structure with Uformer \cite{wang2022uformer} and Restormer \cite{zamir2022restormer} but is composed of different specialized blocks. We ablate the effectiveness of each component in Sec.~\ref{sec:experiment:ablation_study}. 

\subsection{ERFMeter}
\label{sec:method:quantify_erf}

\begin{figure}[t]
    \setlength{\belowcaptionskip}{0.3cm}
	\centering
	\includegraphics[width=\linewidth]{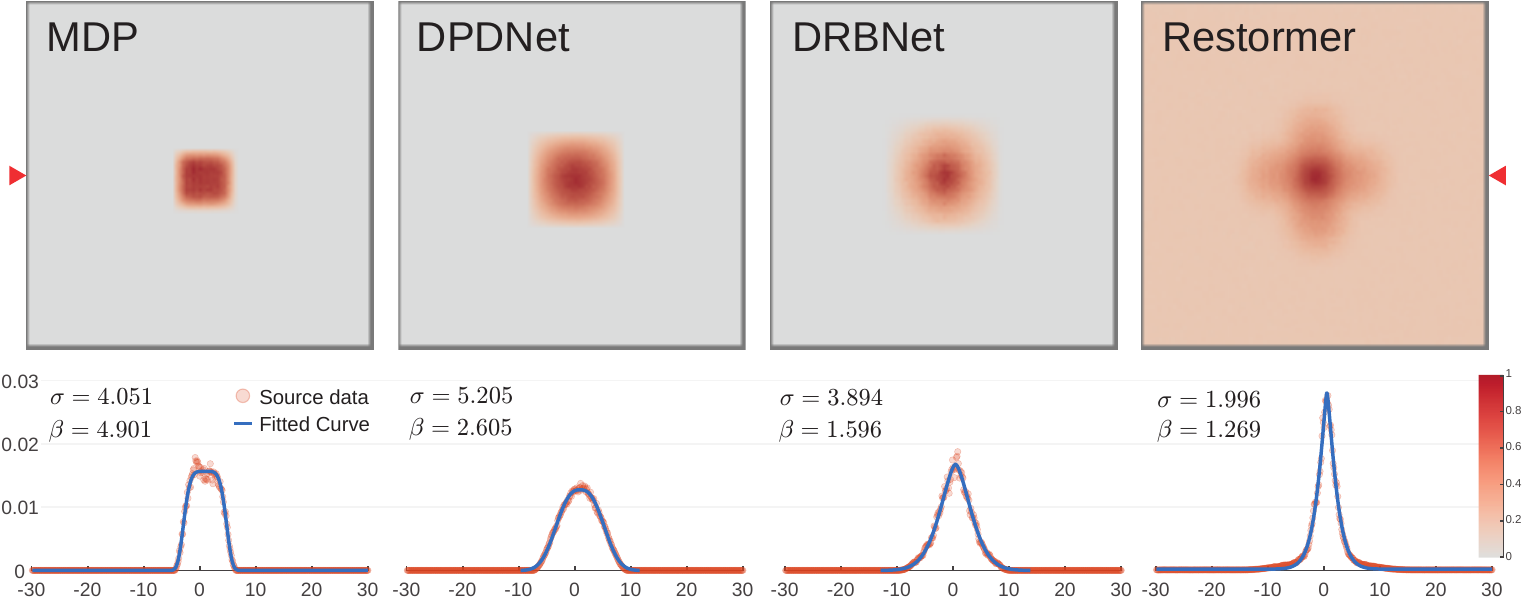}
	\caption{Demonstration of ERF patterns and the fitted GND-PDF curves. Note that, the ERF patterns are in log scale for better visualization, while the GND-PDF curves are in linear scale.}
	\label{fig:erf}

\end{figure}

\begin{figure}[t]
    \setlength{\belowcaptionskip}{0.3cm}
	\centering
	\includegraphics[width=\linewidth]{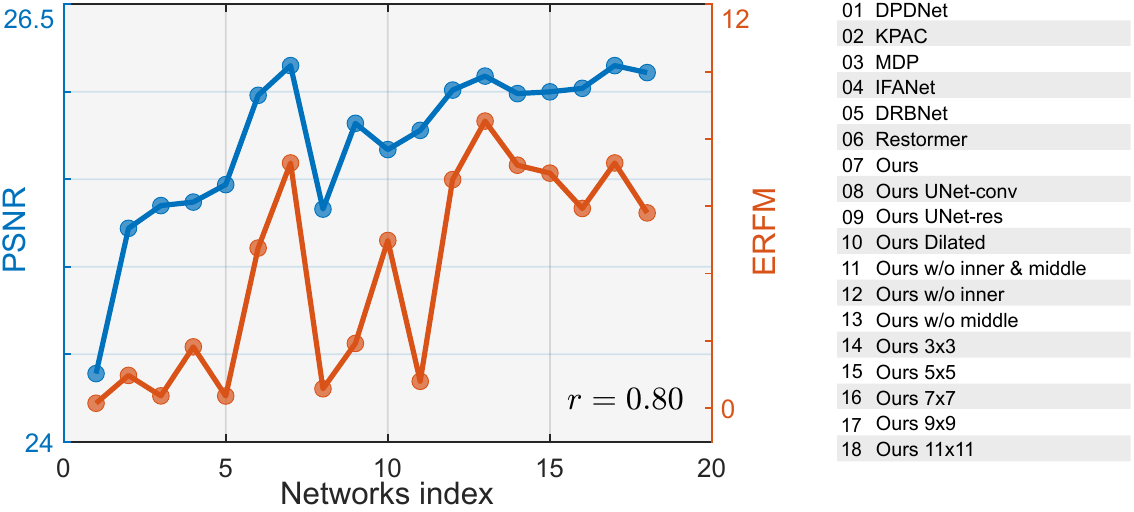}
	\caption{Correlation between ERFM and network performance. All networks were trained and tested on DPDD dataset. The Pearson Correlation Coefficient $r = 0.80$ between ERFM and PSNR of test networks.}
	\label{fig:erf_correlation}

\end{figure}
\begin{figure*}[t]
	\centering
	\includegraphics[width=0.95\linewidth]{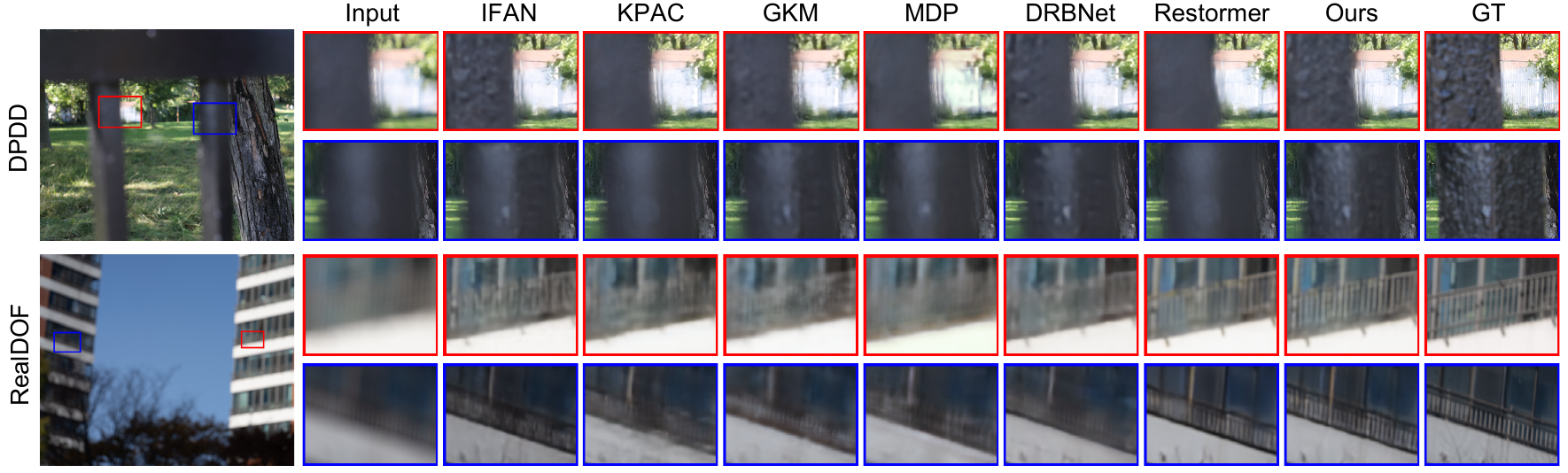}
	\caption{The visual comparison between ours and the recent defocus deblurring methods: IFAN \cite{lee2021iterative}, KPAC \cite{son2021single}, GKM \cite{quan2021gaussian}, Restormer \cite{zamir2022restormer}, MDP \cite{abuolaim2022improving},  DRBNet \cite{ruan2022learning}. Note that DRBNet in this case is trained using only the DPDD dataset. All the methods are evaluated using the code provided by their respective authors. The image samples are taken from the DPDD \cite{abuolaim2020defocus} and RealDOF \cite{lee2021iterative} datasets, respectively. }
	\label{fig:defocus_comparison}
\end{figure*}

In this section, we aim to quantify the influence of ERF. Previous works~\cite{ding2022scaling, liu2022more} mainly focus on enlarging the size of the receptive field through structural re-parameterization\cite{ding2021repvgg} or sparse large kernel. We claim that the size of a receptive field is not the determining factor that accounts for the word ``effective'' while the shape of the receptive field pattern also matters.
We specifically find that the distribution of an ERF can be represented by the Probability Density Function (PDF) of the Generalized Normal Distribution (GND) family. Assuming that the averaged ERF pattern does not depend on the input content, we consider only a symmetric GND that can be derived as: 
\begin{equation}
    f(x) = \frac{c_1\beta}{2\sigma\Gamma\left(\sfrac{1}{\beta}\right)}\exp\left(-\left|\frac{x-\mu}{\sigma}\right|^\beta\right) +c_2
    \label{eq:gnd}
\end{equation}
where $x$ is the raw ERF values, $\mu\in\mathbb{R}$ represents the center of the distribution, and $\sigma\in\mathbb{R}_{>0}$ is the scaling factor that characterizes the variation of distribution: the larger $\sigma$ indicates a more dispersed distribution. The parameter $\beta\in\mathbb{R}_{>0}$ controls the shape of the distribution, \eg $\beta=1$, $\beta=2$, and $\beta\to\infty$ would correspond to the PDF of the Laplace, Gaussian, and Uniform distribution, respectively. The auxiliary parameters $c_1$ and $c_2$ are used to stabilize the curve fitting process and $\Gamma$ is the gamma function that can be calculated as $\Gamma(z) = \int_0^\infty\!x^{z-1}e^{-x}dx$. Fig.~\ref{fig:erf} shows the ERF patterns for existing defocus deblurring networks, along with the fit of $f(x)$ to their central horizontal scanline and the corresponding parameter values. Please refer to the supplementary for more ERFs examples and fitted curves. 

Our mission is to design a network that fully utilizes the information from the entire patch with a large receptive field, and at the same time, focuses more on the adjacent area which has a stronger influence than the peripheral area. These two properties can potentially be  quantified using the estimated parameters of our fitted $f(x)$. 
Specifically, the scaling factor $\sigma$ is used to evaluate the breadth of ERF, where larger $\sigma$ indicates better global attention ability of the network. Moreover, we opt the shape parameter $\beta$ to evaluate the concentration of ERF in which smaller $\beta$ produces sharper distribution in the center and indicates better concentration ability of the network. In summary, we formulate our ERF evaluation method called $\mathsf{ERFMeter}$ as follows:
\begin{equation}
    \mathsf{ERFM} = \frac{\sigma}{\sqrt{2}\beta}\log(\max(x)+1),
    \label{eq:erfmeter}
\end{equation}

Note we also consider the impact of maximum magnitude of the raw ERF response using the scale factor $\log(\max(x)+1)$. We tested our $\mathsf{ERFMeter}$ on several existing defocus deblurring networks and different variants of our own network that were trained and tested on DPDD dataset. As shown in Fig.~\ref{fig:erf_correlation}, the Pearson Correlation coefficient $r=0.80$ indicates the strong correlation between the $\mathsf{ERFM}$ score and network performance (PSNR). While the Eq.~\ref{eq:erfmeter} is just an empirical formula to correlate the network performance with ERF in the image deblurring tasks, we believe that our suggested metric can provide some insights into network structure design and also inspire the research community to investigate similar ERF measures that are suitable for other image-processing tasks. Besides, we only consider a 1D signal (horizontal scanline that crosses the ERF center) for demonstration purposes, the 2D version could be easily deduced.

\section{Experimental Results}
\label{sec:experiment}
\vspace{-0.1cm}
We evaluate our proposed method on the defocus deblurring with single or dual-view images as the input~\cite{abuolaim2020defocus}, as well as single-image motion deblurring task. To evaluate our method in the defocus deblurring task, we train our network on the DPDD dataset, and evaluate it on its own test set and the RealDOF dataset. For motion deblurring, we train our network on the GoPro dataset \cite{nah2017deep}, and evaluate it on its own test set, as well as the HIDE\cite{shen2019human}, RealBlur-J and RealBlur-R \cite{rim2020real} datasets. We additionally train two other models on RealBlur-J and RealBlur-R, respectively.  Some recent works, \eg Restormer-TLC, MPRNet-TLC \cite{chu2021improving} and NAFNet-TLC \cite{chen2022simple}, have been further refined by the test-time improvement tool dubbed TLC \cite{chu2021improving} aiming to narrow the performance gap between the cropped patches and full-resolution images during training and inference. Here we present all the methods without any test time improvement for a fairer comparison. Note that we use the code and trained weights as provided by the respective authors.\\
\noindent \textbf{Implementation Details} \; We adopt AdamW optimizer \cite{loshchilov2017decoupled} with momentum $\beta_{1}=0.9$, $\beta_{2}=0.999$, weight decay 1e-4, learning rate starting from 3e-4 and gradually approach to 1e-6 (cosine annealing \cite{loshchilov2016sgdr}). We follow a similar training strategy as proposed in \cite{zamir2022restormer} where the patch size progressively increases  [192, 256, 288, 368, 448] and the batch size decreases [5, 4, 3, 2, 1] at iterations [60k,  120k, 180k, 240k, 300k] and [180k, 360k, 540k, 720k, 900k] for the defocus and motion deblurring tasks, respectively. For defocus deblurring, we use the charbonnier loss \cite{charbonnier1994two} and perceptual loss \cite{johnson2016perceptual} sequentially for 310K iterations in total, and the charbonnier loss alone for 900k iterations for motion deblurring. More training details regarding specific datasets are included in the supplementary.

\subsection{Comparison to the State-of-the-Art Methods}
\label{sub:compare_to_sotas}

\begin{figure}[ht]
    \setlength{\abovecaptionskip}{0.3cm}
	\centering
	\includegraphics[width=\linewidth]{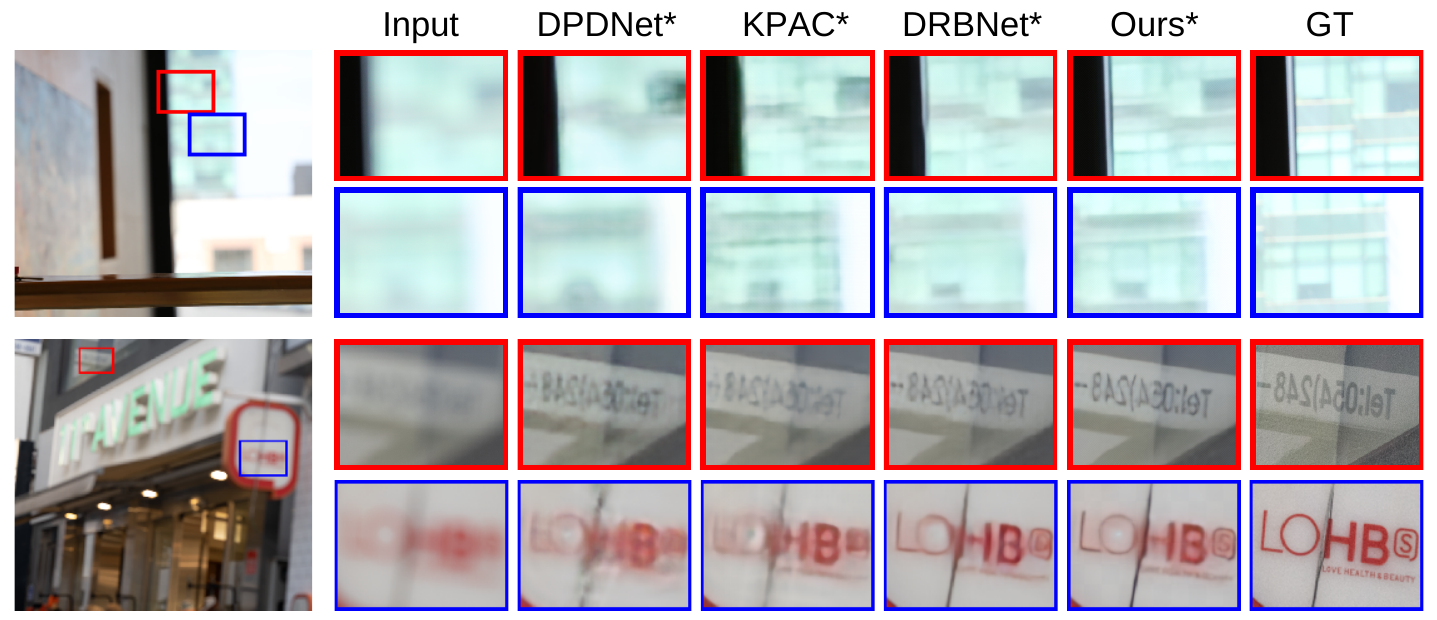}
	\caption{The visual comparison between DPDNet \cite{abuolaim2020defocus}, KPAC \cite{son2021single}, DRBNet \cite{ruan2022learning} and ours when adopting the training strategy in DRBNet \cite{ruan2022learning}. The former three methods are evaluated with the weights and code as provided in  \cite{ruan2022learning}. The sample images are from the DPDD \cite{abuolaim2020defocus} and RealDOF \cite{lee2021iterative} datasets, respectively.}
	\label{fig:performance_gain_lf_dpd}
\end{figure}

\noindent \textbf{Defocus Deblurring} \; We evaluate our method in the defocus deblurring task for the single-image (Tab.~\ref{tab:defocus_comparison_with_others}) and dual-pixel (Tab.~\ref{tab:defocus_comparison_dual}) input. We perform the single-image evaluation on the DPDD \cite{abuolaim2020defocus} and RealDOF \cite{lee2021iterative} datasets, respectively. Specifically, in the former dataset, the defocus and all-in-focus image pairs are captured with wide and narrow apertures individually, while in the latter one by one shot through a customized dual-camera setup with a beam splitter. Note that all compared methods are trained on the DPDD dataset except the AIFNet \cite{ruan2021aifnet} and MDP \cite{abuolaim2022improving} methods, which use their own training set. IFANet \cite{lee2021iterative} adopts extra dual views during the training. Table \ref{tab:defocus_comparison_with_others} demonstrates that our method substantially outperforms existing CNN-based methods, improving over the state-of-the-art method DRBNet \cite{ruan2022learning} by 0.68dB (+2.7\%) PSNR, and slightly outperforms Transformer-based methods, yielding +0.17dB improvement over Restormer \cite{zamir2022restormer}. Notably, our method requires much less computational effort, saving up to 32.2\% parameters and 39\% MACs. Figure \ref{fig:defocus_comparison} further shows that our method is more effective than other approaches qualitatively, particularly, in handling severe defocus (the iron railings) and restoring fine texture (the fence and balcony) in Fig. \ref{fig:defocus_comparison}. The performance of our dual-pixel image defocus deblurring (Tab.~\ref{tab:defocus_comparison_dual}) is similar to its counterpart for the single-image input (Tab.~\ref{tab:defocus_comparison_with_others}), where only for Restormer \cite{zamir2022restormer} the SSIM metric shows slightly better results than our method.  
Additionally, we follow the training strategy that initiates the training on the LFDOF dataset \cite{ruan2021aifnet} followed by the DPDD dataset, showing a better generalization ability than the existing state-of-the-art DRBNet \cite{ruan2022learning}, as illustrated in Tab. \ref{tab:performance_gain_lf_dpd} and Fig. \ref{fig:performance_gain_lf_dpd}. Our method offers a strong restoration ability with sharper details and clearer text.

\begin{table}[htb] \scriptsize
        \caption{Single-image defocus deblurring task: Quantitative image quality and computational cost comparison.}
		\begin{center}
		\resizebox{\linewidth}{!}{
			\begin{tabular}{@{}>{\columncolor{white}[0pt][\tabcolsep]}p{1.5cm}p{0.6cm}p{0.6cm}p{0.6cm}p{0.6cm}p{0.6cm}p{0.6cm}p{0.6cm}>{\columncolor{white}[\tabcolsep][0pt]}p{0.6cm}@{}}
				\toprule
				\multirow{2}{*}[-0.45em]{Method} & \multicolumn{3}{c}{DPDD} & \multicolumn{3}{c}{RealDOF} & \multirow{2}{*}[-0.0em]{\makecell{Params.\\(M))}} & \multirow{2}{*}[-0.0em]{\makecell{MACs\\(G)}}\\
				\cmidrule(rl){2-4}       \cmidrule(l){5-7} 

                 & \multicolumn1c{PSNR$\uparrow$}&\multicolumn1c{SSIM $\uparrow$}&\multicolumn1c{ LPIPS$\downarrow$ }&\multicolumn1c{PSNR$\uparrow$}&\multicolumn1c{SSIM$\uparrow$}&\multicolumn1c{LPIPS$\downarrow$}&\\
				\midrule			
				DPDNet \cite{abuolaim2020defocus} & \hfil 24.39 & \hfil 0.749 &\hfil 0.277   &\hfil  22.87 &\hfil  0.670    &\hfil  0.425 &\hfil 31.03&\hfil 770\\ 
				AIFNet \cite{ruan2021aifnet}        & \hfil 24.21 & \hfil 0.742 &\hfil  0.309  &\hfil  23.09 &\hfil  0.680    &\hfil  0.413 &\hfil 41.55&\hfil 1747\\   
				IFANet\cite{lee2021iterative}     & \hfil 25.37 & \hfil 0.789 &\hfil 0.217   &\hfil  \cellcolor{gray!25}24.71 &\hfil  0.749   &\hfil  0.306  &\hfil 10.48&\hfil 363\\ 
				KPAC \cite{son2021single}     & \hfil 25.22 & \hfil 0.774 &\hfil  0.226   &\hfil  23.98 &\hfil  0.716    &\hfil  0.336  &  \hfil\cellcolor{gray!25}2.06&\hfil  \cellcolor{gray!25}\textbf{113}\\ 
				GKMNet \cite{quan2021gaussian}  & \hfil 25.47 & \hfil 0.786 &\hfil  0.217   &\hfil  24.15 &\hfil  0.728    &\hfil  0.316  &\hfil  \cellcolor{gray!25}\textbf{1.41} &\hfil \cellcolor{gray!25}296 \\ 
				MDP \cite{abuolaim2022improving}  & \hfil 25.35 & \hfil 0.763 &\hfil  0.303   &\hfil  23.73 &\hfil  0.685    &\hfil 0.435  &\hfil 46.86 &\hfil 1898 \\ 	
    		DRBNet \cite{ruan2022learning}       &\hfil 25.47 &\hfil  0.787 &\hfil 0.246  &\hfil  24.70 &\hfil  0.744    &\hfil   0.337  &\hfil 11.69&\hfil 693\\
				Restormer \cite{zamir2022restormer}  & \hfil  \cellcolor{gray!25}{25.98} & \hfil  \cellcolor{gray!25}\textbf{0.811} &\hfil   \cellcolor{gray!25}0.178   &\hfil   \cellcolor{gray!25}\textbf{25.08} &\hfil  \cellcolor{gray!25}\textbf{0.769}    &\hfil \cellcolor{gray!25}0.289  &\hfil 26.13 &\hfil 1983 \\ 				

                \midrule
				Ours        &\hfil \cellcolor{gray!25}\textbf{26.15} &\hfil  \cellcolor{gray!25}{0.810} &\hfil  \cellcolor{gray!25}\textbf{0.155}  &\hfil  \cellcolor{gray!25}\textbf{25.08} &\hfil  \cellcolor{gray!25}{0.762}    &\hfil   \cellcolor{gray!25}\textbf{0.267}  &\hfil 17.7&\hfil 1208\\
				\bottomrule
			\end{tabular}}
		\end{center}
		\label{tab:defocus_comparison_with_others}
\end{table}


 \begin{table}[htb] \scriptsize
  	\caption{Dual-pixel-image defocus deblurring task: Quantitative image quality comparison. Suffix $D$ denotes that the network takes dual-pixel images as the input. }
	\begin{center}
		\begin{tabular}{@{}>{\columncolor{white}[0pt][\tabcolsep]}p{2.2cm}p{0.8cm}p{0.8cm}p{0.8cm}p{0.8cm}>{\columncolor{white}[\tabcolsep][0pt]}p{0.8cm}@{}}
		\toprule
		\multirow{2}{*}[-0.45em]{ Method} & \multicolumn{3}{c}{DPDD} & \multirow{2}{*}[-0.0em]{\makecell{Params.\\(M)}}  \\ 
			\cmidrule(rl){2-4}                      
			 & PSNR$\uparrow$  & SSIM$\uparrow$ & LPIPS$\downarrow$    \\ 
			\midrule
			
			 DPDNet$_D$ \cite{abuolaim2020defocus}  &  25.13  & 0.786 & 0.223 & \hfil 31.03   \\   
              IFAN$_D$ \cite{lee2021iterative}  &  25.99  & 0.804 & 0.207 & \hfil 10.48    \\    
              KPAC$_D$ \cite{son2021single} &  25.82  & 0.800 & 0.185 &\hfil \cellcolor{gray!25}2.06    \\    
              RDPD$_D$ \cite{abuolaim2021learning} &  25.41  & 0.771 & 0.255 &\hfil \cellcolor{gray!25}\textbf{1.41}     \\  
            DRBNet$_D$ \cite{ruan2022learning} &  26.33  & 0.811 & \cellcolor{gray!25}{0.154} &\hfil 11.69   \\ 
            Restormer$_D$ \cite{zamir2022restormer} &  \cellcolor{gray!25}{26.66}  & \cellcolor{gray!25}\textbf{0.833} & 0.155 &\hfil 26.13     \\ 
              \midrule
               Ours$_D$  &  \cellcolor{gray!25}\textbf{26.72}  & \cellcolor{gray!25}{0.826} & \cellcolor{gray!25}\textbf{0.140} & \hfil 17.7  \\       
			\bottomrule
		\end{tabular}
	\end{center}
	\label{tab:defocus_comparison_dual}
\end{table}


	\begin{table} \scriptsize 
		\setlength{\abovecaptionskip}{0.3cm}
		\caption{The performance comparison among DPDNet \cite{abuolaim2020defocus}, KPAC \cite{son2021single}, DRBNet \cite{ruan2022learning} and our method when trained on LFDOF \& DPDD. * denotes that each respective network is trained with the strategy as proposed in \cite{ruan2022learning}. }
		\begin{center}
			\begin{tabular}{@{}>{\columncolor{white}[0pt][\tabcolsep]}p{1.6cm}p{0.6cm}p{0.6cm}p{0.6cm}p{0.6cm}p{0.6cm}p{0.6cm}}
				\toprule
				\multirow{2}{*}[-0.45em]{ Method} & \multicolumn{3}{c}{DPDD} & \multicolumn{3}{c}{RealDOF} \\ 
				\cmidrule(rl){2-4}                      \cmidrule(l){5-7}
				& PSNR$\uparrow$  & SSIM$\uparrow$ & LPIPS$\downarrow$  & PSNR$\uparrow$  & SSIM$\uparrow$ & LPIPS$\downarrow$  \\ 
				\midrule
				 
				DPDNet* & 24.90  & 0.761 & 0.278  & 24.16  & 0.712 & 0.377   \\ 
				KPAC* &  25.47  & 0.780 & 0.220 & 24.64  & 0.735 & 0.319  \\
				DRBNet* & \cellcolor{gray!25}{25.73} & \cellcolor{gray!25}{0.791}& \cellcolor{gray!25}{0.183}   & \cellcolor{gray!25}{25.75} & \cellcolor{gray!25}\textbf{0.771}& \cellcolor{gray!25}{0.257}   \\     
				Ours* & \cellcolor{gray!25}\textbf{25.89} & \cellcolor{gray!25}\textbf{0.792}& \cellcolor{gray!25}\textbf{0.154}   & \cellcolor{gray!25}\textbf{25.83} & \cellcolor{gray!25}{0.769}& \cellcolor{gray!25}\textbf{0.245}   \\    
				\bottomrule
			\end{tabular}
		\end{center}

		\label{tab:performance_gain_lf_dpd}
	\end{table}

\begin{figure*}[t]
	\centering
	\includegraphics[width=0.97\linewidth]{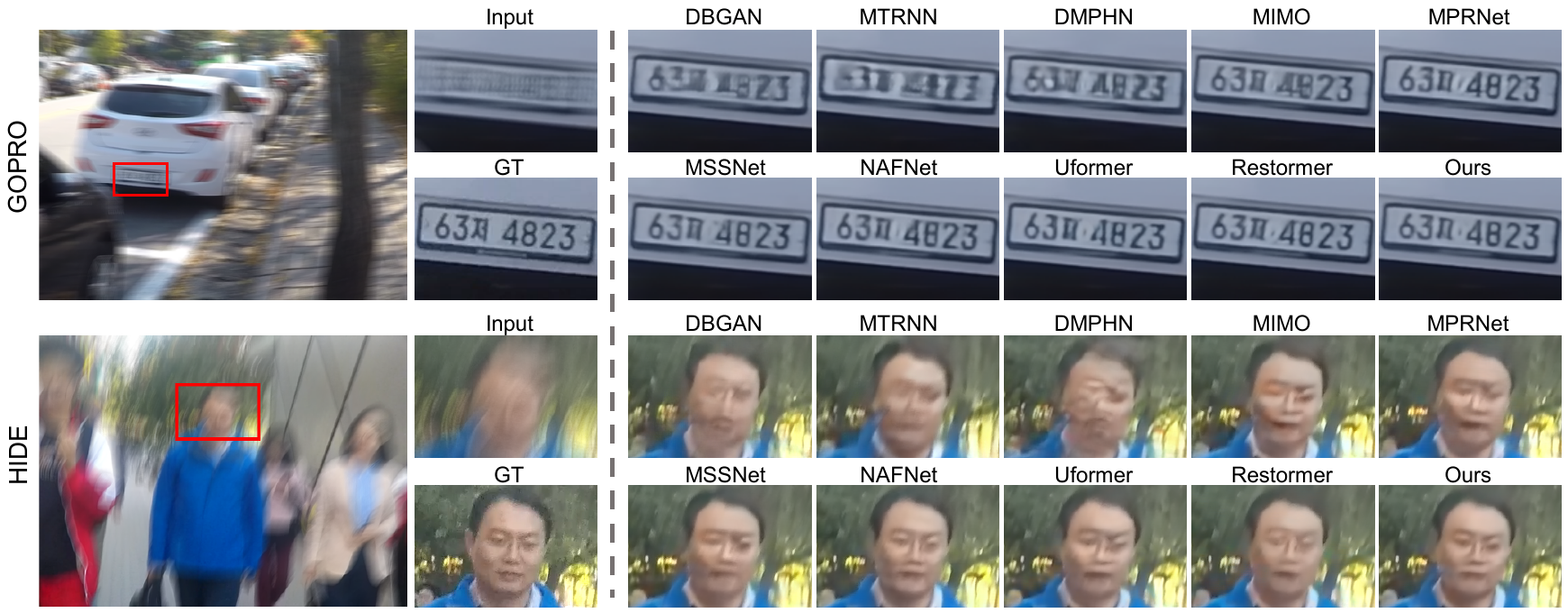}
	\caption{The visual comparison between our method and the recent motion deblurring techniques including DBGAN \cite{zhang2020deblurring}, MTRNN \cite{park2020multi}, DMPHN \cite{zhang2019deep}, MIMO \cite{cho2021rethinking}, MPRNet \cite{zamir2021multi}, MSSNet \cite{kim2022mssnet}, NAFNet \cite{chen2022simple}, Uformer \cite{wang2022uformer}, and Restormer \cite{zamir2022restormer}. All the methods are evaluated using the code provided by their respective authors without any test-time refinement (\eg TLC \cite{chu2021improving}) for fairer comparisons. The upper and bottom row images are taken from the GOPRO \cite{nah2017deep} and HIDE \cite{shen2019human} datasets, respectively.  }
	\label{fig:motion_comparison}
\end{figure*}

\paragraph{Motion deblurring} \; We evaluate the performance of our network for single-image motion deblurring using four benchmark datasets with synthetic blur (GoPro \cite{nah2017deep}, HIDE \cite{shen2019human}) and real-world blur (RealBlur-J and RealBlur-R \cite{rim2020real}).
We compare our network performance with the following state-of-the-art learning-based techniques:
CNNs (\cite{nah2017deep}, \cite{zhang2019deep}, \cite{tao2018scale}, \cite{shen2019human}, \cite{gao2019dynamic}, \cite{zhang2019deep}, \cite{suin2020spatially}, \cite{gao2019dynamic}, \cite{purohit2021spatially}, \cite{cho2021rethinking}), GANs (\cite{kupyn2018deblurgan}, \cite{kupyn2019deblurgan}, \cite{zhang2020deblurring}), RNNs (\cite{park2020multi}), and Transformers (\cite{wang2022uformer}, \cite{zamir2022restormer}). 
First, following \cite{wang2022uformer,zamir2022restormer} we evaluate our method on all four datasets while training on the GoPro dataset alone.
Table \ref{tab:comparison_with_others_motion_gopro} shows that we achieve a competitive performance on the GoPro and HIDE datasets, significantly surpassing the existing CNN, GAN, and RNN solutions. For instance, we outperform the latest MSSNet \cite{kim2022mssnet} and NAFNet \cite{chen2022simple} (CNN) by +0.34dB / +0.27dB on the GoPro in terms of PSNR. Our method also outperforms Transformer-based Uformer / Restormer by +0.30dB / +0.43dB on the GoPro dataset, while sparing up to 66.4\% / 34.6\% parameters and 47.5\% / 43.3\% MACs. We use gray background to indicate the top two competitors and extra bold font for the champion. Second, we additionally evaluate our network on the datasets with real-world blur \cite{rim2020real} as shown in Tab.~\ref{tab:motion_comparison_R_J}. The upper part presents the outcome of training on the GoPro dataset with synthetic blur.  Two Transformer-based solutions perform slightly better than ours when it comes to generalizing from synthetic to real blur. This is expected due to our compact structure, while we still obtain comparable or even better results than the remaining competitors, for example, requiring 89.7\% and 47.9\% fewer MACs compared to MPRNet and MSSNet. The lower part of Tab.~\ref{tab:motion_comparison_R_J} refers to respective training on the RealBlur-J and RealBlur-R datasets in which case our network performs the best with +0.60dB / +0.15dB and +0.57dB /+ 0.23dB gain over MPRNet / MSSNet. Figure \ref{fig:motion_comparison} demonstrates how our method can restore challenging examples with a plausible visual quality when compared to others (\eg text and face). 

\begin{table}[htb] \scriptsize
	\caption{Motion deblurring comparison on the GoPro dataset. Please note the reported NAFNet is without TLC for fair  network structure benchmark comparison. Please refer to its TLC version in \cite{chen2022simple}.}   
	\begin{center}
        \resizebox{\columnwidth}{!}{
		\begin{tabular}{@{}>{\columncolor{white}[0pt][\tabcolsep]}p{2.1cm}p{0.6cm}p{0.6cm}p{0.6cm}p{0.6cm}p{0.7cm}>{\columncolor{white}[\tabcolsep][0pt]}p{0.7cm}@{}}
			\toprule
			\multirow{2}{*}[-0.45em]{Method}  & \multicolumn{2}{c}{GoPro} & \multicolumn{2}{c}{HIDE}  &\multirow{2}{*}[-0.0em]{\makecell{Params.\\ (M)}} &\multirow{2}{*}{\makecell{MACs\\(G)}}\\
			\cmidrule(l){2-3}                      \cmidrule(l){4-5}  
                & \multicolumn1c{PSNR$\uparrow$}&\multicolumn1c{ SSIM$\uparrow$}&\multicolumn1c{PSNR$\uparrow$}&\multicolumn1c{SSIM$\uparrow$}&\\
			\midrule   
			 DeblurGAN \cite{kupyn2018deblurgan}  & \hfil 28.70 & \hfil 0.858 &\hfil 24.51   &\hfil  0.871   &\hfil 6.06  &\hfil 809 \\ 
			 DeepDeblur \cite{nah2017deep} & \hfil 29.08 & \hfil 0.914 &\hfil 25.73   &\hfil  0.874 &\hfil  11.72 &\hfil 4729    \\ 
			 Zhang \etal \cite{zhang2018dynamic}   & \hfil 29.19 & \hfil 0.931 &\hfil N/A  &\hfil  N/A  &\hfil 37.1 &\hfil N/A  \\ 
			 DeblurGAN-v2 \cite{kupyn2019deblurgan}  & \hfil 29.55 & \hfil 0.934 &\hfil 26.61   &\hfil  0.875  &\hfil 5.08  &\hfil \cellcolor{gray!25}\textbf{411}\\ 
			 SRN \cite{tao2018scale}   & \hfil 30.26 & \hfil 0.934 &\hfil 28.36   &\hfil  0.915  &\hfil  8.06 &\hfil 20134 \\ 
			 Shen \etal \cite{shen2019human}   & \hfil 30.26 &\hfil 0.940 &\hfil 28.89   &\hfil  0.930  &\hfil N/A  &\hfil N/A\\ 
    	Gao \etal \cite{gao2019dynamic}   & \hfil 30.90 & \hfil 0.935 &\hfil 29.11   &\hfil  0.913 &\hfil  \cellcolor{gray!25}2.84  &\hfil 3255     \\ 
        DBGAN \cite{zhang2020deblurring} & \hfil 31.10 & \hfil 0.942 &\hfil 28.94   &\hfil  0.915 &\hfil  11.59   &\hfil 10685   \\ 
        MTRNN \cite{park2020multi}  & \hfil 31.15 & \hfil 0.945 &\hfil 29.15  &\hfil  0.918 &\hfil  \cellcolor{gray!25}\textbf{2.6}  &\hfil 2315 \\ 
        DMPHN \cite{zhang2019deep}   & \hfil 31.20 & \hfil 0.940 &\hfil 29.09   &\hfil  0.924 &\hfil 7.23 &\hfil 1100    \\ 
        Suin \etal \cite{suin2020spatially}   & \hfil 31.85 & \hfil 0.948 &\hfil 29.98   &\hfil  0.930 &\hfil  N/A &\hfil  N/A    \\ 
        SPAIR \cite{purohit2021spatially}   & \hfil 32.06 & \hfil 0.953 &\hfil30.29   &\hfil  0.931&\hfil  N/A &\hfil  N/A  \\ 
        MIMO-UNet+ \cite{cho2021rethinking}   & \hfil 32.45 & \hfil 0.957 &\hfil29.99   &\hfil  0.930&\hfil  16.1 &\hfil 2171  \\ 
        MPRNet \cite{zamir2021multi} & \hfil 32.66 & \hfil 0.959  &\hfil  30.96 &\hfil 0.939  &\hfil 20.1 &\hfil 10927 \\ 
        Uformer \cite{wang2022uformer}   & \hfil 33.05 & \hfil 0.962 &\hfil 30.89   &\hfil  0.940 & \hfil 50.88 & \hfil 2143   \\ 
        Restormer \cite{zamir2022restormer}   & \hfil 32.92 & \hfil 0.961 &\hfil \cellcolor{gray!25}\textbf{31.22}   &\hfil  \cellcolor{gray!25}{0.942} &\hfil 26.13 &\hfil 1983\\ 
        MSSNet \cite{kim2022mssnet} & \hfil 33.01 & \hfil 0.961  &\hfil  30.79 &\hfil 0.938  &\hfil 15.59 &\hfil 2159 \\ 
        NAFNet \cite{chen2022simple} & \hfil \cellcolor{gray!25}{33.08} & \hfil \cellcolor{gray!25}{0.963} & \hfil \cellcolor{gray!25}\textbf{31.22} &\hfil \cellcolor{gray!25}\textbf{0.943}  &\hfil  67.89  &\hfil  \cellcolor{gray!25}890  \\ 
        
\midrule
        Ours   \hfil &  \cellcolor{gray!25}\textbf{33.35} & \hfil \cellcolor{gray!25}\textbf{0.964} &\hfil \cellcolor{gray!25}{31.21}   &\hfil  \cellcolor{gray!25}\textbf{0.943}  &\hfil 17.1 &\hfil 1125 \\ 
			\bottomrule
		\end{tabular}}
	\end{center}

	\label{tab:comparison_with_others_motion_gopro}
\end{table}

\begin{table}[htb] \scriptsize
  	\caption{Motion deblurring comparison on the RealBlur-R and RealBlur-J datasets. The upper part refers to training restricted to the GoPro dataset with synthesized blur, while in the lower part training is performed using the individual datasets with real-world blur. Please note the reported NAFNet is without TLC for fair  network structure benchmark comparison. Please refer to its TLC version in \cite{chen2022simple}.}
	\begin{center}
        \resizebox{\columnwidth}{!}{
		\begin{tabular}{@{}>{\columncolor{white}[0pt][\tabcolsep]}p{2.25cm}p{0.6cm}p{0.6cm}p{0.6cm}p{0.6cm}p{0.6cm}p{0.6cm}p{0.7cm}>{\columncolor{white}[\tabcolsep][0pt]}p{0.7cm}@{}}
		\toprule
		\multirow{2}{*}[-0.45em]{ Method} & \multicolumn{2}{c}{RealBlur-R} & \multicolumn{2}{c}{RealBlur-J} & \multirow{2}{*}[-0.0em]{\makecell{Params.\\(M)}} & \multirow{2}{*}[-0.0em]{\makecell{MACs\\(G)}}\\ 
			\cmidrule(rl){2-3}                      \cmidrule(l){4-5}
			 & PSNR$\uparrow$  & SSIM$\uparrow$  & PSNR$\uparrow$  & SSIM$\uparrow$   \\ 
			\midrule

			 Hu et al. \cite{hu2014deblurring} &  33.67  & 0.916 & 26.41 & 0.803  & \hfil N/A & \hfil N/A  \\   
    DeepDeblur \cite{nah2017deep} &  32.51  & 0.841 & 27.87 & 0.827 &  \hfil 11.72 & \hfil 4279  \\    
    DeblurGAN \cite{kupyn2018deblurgan} &  33.79  & 0.903 & 27.97 & 0.834  & \hfil \cellcolor{gray!25}6.06  & \hfil \cellcolor{gray!25}809  \\    
    Pan \etal \cite{pan2016blind} &  34.01  & 0.916 & 27.22 & 0.790 & \hfil N/A & \hfil N/A   \\
    DeblurGAN-v2 \cite{kupyn2019deblurgan} &  35.26  & 0.944 & 28.70 & 0.866  & \hfil 5.08  & \hfil 411   \\
    Zhang \etal \cite{zhang2018dynamic} &  35.48  & 0.947 & 27.80 & 0.866 &\hfil 37.1 &\hfil N/A    \\
    SRN \cite{tao2018scale} &  35.66  & 0.947 & 28.56 & 0.867 &\hfil 8.06   & 20134\\
    DMPHN \cite{zhang2019deep} &  35.70  & 0.948 & 28.42 & 0.860 &\hfil 7.23 &\hfil 1100   \\
    MPRNet \cite{zamir2021multi} &  35.99  & 0.952 & 28.70 & 0.873 &\hfil 20.1 & 10927   \\
    MSSNet \cite{kim2022mssnet} &  35.93  & 0.953 & 28.79 & 0.879 &\hfil 15.6 &\hfil 2159   \\
    
    Uformer \cite{wang2022uformer} &  36.22  & 0.957 & 29.06 & 0.884 & \hfil 50.88 & \hfil 2143 \\
    Restormer \cite{zamir2022restormer} & 36.19  & 0.957 & 28.96 & 0.879 &\hfil 26.13 &\hfil 1983 \\
    NAFNet \cite{chen2022simple} &  36.14 &  0.955 &  28.43 & 0.860  &\hfil  67.89  &\hfil  890  \\ 
    \midrule
        Ours &  35.91 & 0.954 & 28.78 & 0.878 &\hfil 17.1 &\hfil 1125   \\
    \midrule
    \midrule
    DeblurGAN-v2 \cite{kupyn2019deblurgan} &  36.44  & 0.935 & 29.69 & 0.870 &\hfil \cellcolor{gray!25}\textbf{5.08}  &\hfil \cellcolor{gray!25}\textbf{411}   \\   
    SRN \cite{tao2018scale} &  38.65  & 0.965 & 31.38 & 0.909 &\hfil 8.06   & 20134   \\    
    MPRNet \cite{zamir2021multi} &  39.31  & \cellcolor{gray!25}{0.972} & 31.76 & 0.922  &\hfil 20.1 & 10927  \\    
    MIMO-UNet++ \cite{cho2021rethinking} &\hfil  N/A  &\hfil N/A & 32.05 & 0.921 &\hfil 16.1 &\hfil 8683   \\   
    MSSNet \cite{kim2022mssnet} &  \cellcolor{gray!25}{39.76}  & \cellcolor{gray!25}{0.972} & \cellcolor{gray!25}{32.10} & \cellcolor{gray!25}{0.928} &\hfil 15.6  &\hfil 2159  \\ 
    \midrule
    Ours &  \cellcolor{gray!25}\textbf{39.91}  & \cellcolor{gray!25}\textbf{0.974} & \cellcolor{gray!25}\textbf{32.33} & \cellcolor{gray!25}\textbf{0.929} &\hfil 17.1 &\hfil 1125 \\     
			\bottomrule
		\end{tabular}}
	\end{center}
	\label{tab:motion_comparison_R_J}
\end{table}

\subsection{Ablation study}
\label{sec:experiment:ablation_study}
We conduct our ablation study on the DPDD dataset for defocus debluring and the GoPro dataset for motion deblurring. We start by explaining the effectiveness of our proposed LaKD block compared to the existing UNet baseline (pure convolution and residual blocks) and then compare it to an equivalent block with dilated convolution. We further discuss the key design decisions concerning the shortcuts, kernel size, and block number. The relevant network structure could be found in the supplementary.

\begin{table}[htb] \scriptsize
        \caption{Quantitative performance of three UNet-like structures.}
	\begin{center}
		\begin{tabular}{@{}>{\columncolor{white}[0pt][\tabcolsep]}p{1.8cm}p{0.8cm}p{0.8cm}>{\columncolor{white}[\tabcolsep][0pt]}p{0.8cm}@{}}
		\toprule
		\multirow{2}{*}[-0.45em]{ Method} & \multicolumn{3}{c}{DPDD}  \\ 
			\cmidrule(rl){2-4}                      
			 & PSNR$\uparrow$  & SSIM$\uparrow$ & LPIPS$\downarrow$    \\ 
			\midrule
			
			 UNet-conv \cite{ronneberger2015u} &  25.33  & 0.775 & 0.216   \\    
              UNet-res \cite{lim2017enhanced} &  25.82  & 0.797 & 0.178   \\   
              Ours  &  \cellcolor{gray!25}\textbf{26.15}  & \cellcolor{gray!25}\textbf{0.810} & \cellcolor{gray!25}\textbf{0.155}     \\    
    
			\bottomrule
		\end{tabular}
	\end{center}

	\label{tab:baseline_comparison}
\end{table}



\noindent \textbf{Effectiveness of LaKD block} \; We adopt two  UNet-like CNNs composed of pure convolution layers \cite{ronneberger2015u} and Resblock \cite{lim2017enhanced}, denoted as UNet-conv and UNet-res. Table \ref{tab:baseline_comparison} demonstrates that our proposed block largely outperforms these two baselines by 3.2\%  (+0.82dB) and 1.3\% (+0.33dB). This demonstrates the superiority of our customized CNN block. \\
\noindent \textbf{Large kernel \vs Dilated convolution} \; We replace our large kernel depth-wise convolution with dilated (atrous) convolutions \cite{yu2015multi}, which could also expand the receptive field \cite{seif2018large} or capture long-range information \cite{wu2016bridging} through defining spacing between the convolutional filters with various dilation rates. We adopt the hybrid dilated convolution (HDC) to avoid gridding artifacts inherited from a serialized convolution with a fixed dilation rate \cite{wang2018understanding}. Here we specify the dilation rates of 1, 2, 3 in the three cascaded convolution layers accordingly inside the block and preserve the equivalent inner and middle shortcuts.
Table \ref{tab:dilated_ours} shows that our LaKD block equipped with large depth-wise convolution significantly outperforms dilated convolution in both defocus and motion deblurring tasks. 
This can be explained by the nature of dilated convolution, where the local or neighboring information is lost due to the sparsity in their convolution kernels.

\begin{table}[htb] \scriptsize
  	\caption{Quantitative performance comparison between the model with dilated convolution and our LaKDNet.}
	\begin{center}
		\begin{tabular}{@{}p{0.6cm}p{0.6cm}p{0.6cm}p{0.6cm}p{0.8cm}p{0.6cm}p{0.6cm}p{0.8cm}}
		\toprule
		\multirow{2}{*}[-0.45em]{ Method} & \multicolumn{3}{c}{Defocus} &\multirow{2}{*}[-0.0em]{\makecell{Params.\\(M)}} & \multicolumn{2}{c}{Motion} &\multirow{2}{*}[-0.0em]{\makecell{Params.\\(M)}}  \\ 
			\cmidrule(rl){2-4}                      \cmidrule(l){6-7}
			 & PSNR$\uparrow$  & SSIM$\uparrow$ & LPIPS$\downarrow$ & & PSNR$\uparrow$  & SSIM$\uparrow$  \\ 
			\midrule
			
			 Dilated & \hfil25.67  & \hfil0.785 & \hfil0.195 & \hfil21.9 & \hfil32.82  & \hfil0.960 & \hfil21.9  \\   
			 Ours &  \hfil\cellcolor{gray!25}\textbf{26.15}  & \hfil\cellcolor{gray!25}\textbf{0.810} & \hfil\cellcolor{gray!25}\textbf{0.155} & \hfil\cellcolor{gray!25}\textbf{17.7} & \hfil\cellcolor{gray!25}\textbf{33.35}  & \hfil\cellcolor{gray!25}\textbf{0.962}  & \hfil\cellcolor{gray!25}\textbf{17.1}  \\

			\bottomrule
		\end{tabular}
	\end{center}
	\label{tab:dilated_ours}
\end{table}

\noindent \textbf{Importance of mixing} \; The shortcut in our LaKD module offers considerable improvement to network performance. As shown in Tab.~\ref{tab:skip_connection}, we ablate three variants of the shortcuts: (1) removing the inner shortcut inside the feature mixing module (labeled in red in Fig. \ref{fig:network}); (2) removing the middle shortcut starting from the beginning of the LaKD block till the end of feature mixing module (labeled in blue in Fig. \ref{fig:network}); (3) removing the inner and middle shortcuts. Note that we always keep the initial shortcut from the beginning to the end of the LaKD block (labeled in black in Fig. \ref{fig:network}) as it is helpful for the final performance as demonstrated in UNet-res (Tab. \ref{tab:baseline_comparison}). Table \ref{tab:skip_connection} shows that the performance will drop significantly without inner and middle shortcuts, even slightly worse than the baseline UNet-res (Tab. \ref{tab:baseline_comparison}). This is expected since we use unusually large kernels for training, where shortcuts could potentially benefit the gradient flow and feature propagation.  
\begin{table}[htb] \scriptsize
  	\caption{The effectiveness of shortcuts inside the LaKD block.}
	\begin{center}
		\begin{tabular}{@{}>{\columncolor{white}[0pt][\tabcolsep]}p{2.4cm}p{1.1cm}p{1.1cm}>{\columncolor{white}[\tabcolsep][0pt]}p{1.1cm}@{}}
		\toprule
		\multirow{2}{*}[-0.45em]{ Method} & \multicolumn{3}{c}{DPDD}  \\ 
			\cmidrule(rl){2-4}                      
			 & PSNR$\uparrow$  & SSIM$\uparrow$ & LPIPS$\downarrow$    \\ 
			\midrule
			
			 w/o inner &  26.01  & 0.807 & 0.156   \\   
              w/o middle  &  26.09  & \cellcolor{gray!25}\textbf{0.810} & 0.159    \\ 
              w/o inner \& middle  &  25.78  & 0.795 & 0.175   \\ 
              Ours  &  \cellcolor{gray!25}\textbf{26.15}  & \cellcolor{gray!25}\textbf{0.810} & \cellcolor{gray!25}\textbf{0.155}    \\     
			\bottomrule
		\end{tabular}
	\end{center}
	\label{tab:skip_connection}
\end{table}

\noindent \textbf{Small or large kernel size?} \; Table \ref{tab:kernel_size} shows the influence of various kernel sizes on the performance. 
Similar as ~\cite{liu2022convnet}, we find the performance saturated on kernel size of $9 \times 9$ for defocus and $7 \times 7$ for motion deblurring. Tricks like structural reparameterization~\cite{ding2021repvgg}, kernel decomposition or dynamic sparsity~\cite{liu2022more} could possibility be applied to further enlarge the ERF without performance dropping in the future. Besides, it is worth noting that our LaKD is able to achieve a relatively larger ERF, despite using a common small kernel size of $3 \times 3$, as shown in Fig.~\ref{fig:erf} and Fig.~\ref{fig:erf_kernel}. When applying the kernel size larger than $5 \times 5$, the receptive field is covering the whole $512\times512$ patch.
\begin{table}[htb] \scriptsize
  	\caption{Quantitative performance with respect to kernel size.}
	\begin{center}
        \resizebox{\columnwidth}{!}{
		\begin{tabular}{@{}>{\columncolor{white}[0pt][\tabcolsep]}p{1.0cm}p{0.6cm}p{0.65cm}p{0.6cm}p{0.8cm}p{0.6cm}p{0.6cm}>{\columncolor{white}[\tabcolsep][0pt]}p{0.8cm}@{}}
		\toprule
		\multirow{2}{*}[-0.45em]{ Method} & \multicolumn{3}{c}{Defocus} &\multirow{2}{*}[-0.0em]{\makecell{Params.\\(M)}} & \multicolumn{2}{c}{Motion} &\multirow{2}{*}[-0.0em]{\makecell{Params.\\(M)}}  \\ 
			\cmidrule(rl){2-4}                      \cmidrule(l){6-7}
			 & PSNR$\uparrow$  & SSIM$\uparrow$ & LPIPS$\downarrow$ & & PSNR$\uparrow$  & SSIM$\uparrow$  \\ 
			\midrule
			
			\hfil $3 \times 3$  &  \hfil25.99  & \hfil0.809 & \hfil0.162 & \hfil\cellcolor{gray!25}\textbf{16.4} & \hfil33.21  & \hfil0.964 & \hfil\cellcolor{gray!25}\textbf{16.4}   \\   
			 \hfil $5 \times 5$ &  \hfil26.00  & \hfil0.809 & \hfil0.163 & \hfil16.7 & \hfil33.26  & \hfil0.964  & \hfil16.7  \\ 
			\hfil  $7 \times 7$ &  \hfil26.02  & \hfil0.805 & \hfil0.155 & \hfil17.1 & \hfil\cellcolor{gray!25}\textbf{33.35}  & \hfil\cellcolor{gray!25}\textbf{0.964} & \hfil17.1   \\  
    	    \hfil $9 \times 9$ &  \hfil\cellcolor{gray!25}\textbf{26.15}  & \hfil\cellcolor{gray!25}\textbf{0.810} & \hfil0.155 & \hfil17.7& \hfil33.30  & \hfil0.964 & \hfil17.7 \\ 
    	   \hfil  $11 \times 11$ &  \hfil26.11  & \hfil0.806 & \hfil\cellcolor{gray!25}\textbf{0.154} & \hfil18.5& \hfil N/A & \hfil N/A & \hfil N/A \\ 			       
			\bottomrule
		\end{tabular}}
	\end{center}
	\label{tab:kernel_size}
\end{table}

\begin{figure}[t]
	\centering
	\includegraphics[width=\linewidth]{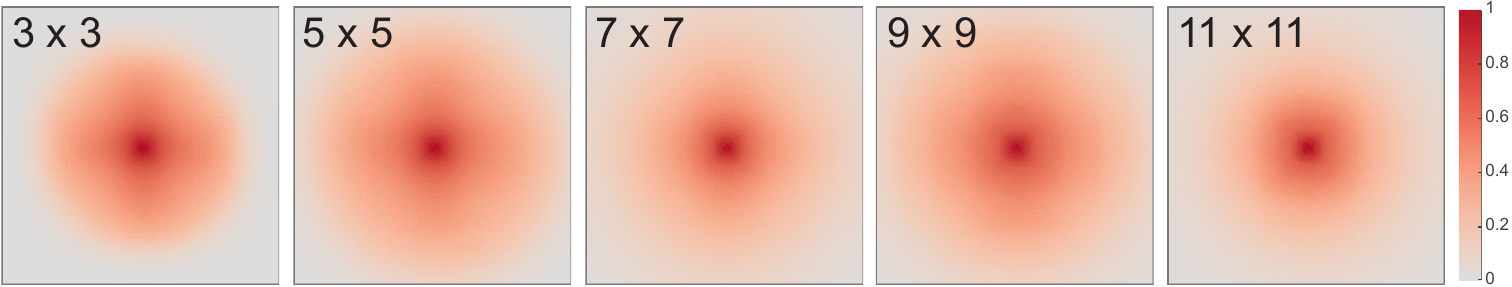}
	\caption{ERF patterns (DPDD dataset) when using different kernel sizes in our network.}
	\label{fig:erf_kernel}

\end{figure}

\noindent \textbf{Effect of block number} \; Table \ref{tab:block_num} shows that the performance increases with the network scale, where the first column indicates the block number $[N1, N2, N3, N4]$ in the 4-level symmetric encoder-decoder. A larger scale network generally results in better performance but comes at an extra cost. For defocus deblurring, $[10,14,14,18]$ achieves 0.2\% (+0.05 dB) PSNR improvement with 12.9\% (+2.2 M) parameter increase than $[8,12,12,16]$, while for motion deblurring, we get 0.14dB PSNR drop. We, therefore, choose $[8,12,12,16]$ as our final setting considering the trade-off on scale and performance.  
\begin{table}[htb] \scriptsize
  	\caption{Quantitative performance with respect to block numbers.}
	\begin{center}
         \resizebox{\columnwidth}{!}{
		\begin{tabular}{@{}>{\columncolor{white}[0pt][\tabcolsep]}p{1.5cm}p{0.6cm}p{0.6cm}p{0.6cm}p{0.8cm}p{0.6cm}p{0.6cm}>{\columncolor{white}[\tabcolsep][0pt]}p{0.8cm}@{}}
		\toprule
		\multirow{2}{*}[-0.45em]{ Method} & \multicolumn{3}{c}{Defocus} &\multirow{2}{*}[-0.0em]{\makecell{Params.\\(M)}} & \multicolumn{2}{c}{Motion} &\multirow{2}{*}[-0.0em]{\makecell{Params.\\(M)}}  \\ 
			\cmidrule(rl){2-4}                      \cmidrule(l){6-7}
			 & PSNR$\uparrow$  & SSIM$\uparrow$ & LPIPS$\downarrow$ & & PSNR$\uparrow$  & SSIM$\uparrow$  \\ 
			\midrule
			
			 $[4, 6, 6, 8]$ &  \hfil25.84  & \hfil0.795 & \hfil0.166 & \hfil\cellcolor{gray!25}\textbf{9.7}  & \hfil32.86  & \hfil0.961 & \hfil\cellcolor{gray!25}\textbf{9.4}   \\ 
    	   $[6, 8, 8, 10]$ &  \hfil25.90  & \hfil0.788 & \hfil0.162 & \hfil12 & \hfil33.09  & \hfil0.962  & \hfil11.6  \\ 
			 $[8, 12, 12, 16]$ &  \hfil26.15  & \hfil0.810 & \hfil0.155 & \hfil17.7 & \hfil\cellcolor{gray!25}\textbf{33.35}  & \hfil\cellcolor{gray!25}\textbf{0.964} & \hfil17.1   \\  
    	   $[10, 14, 14, 18]$ &  \hfil\cellcolor{gray!25}\textbf{26.20}  & \hfil\cellcolor{gray!25}\textbf{0.812} & \hfil\cellcolor{gray!25}\textbf{0.151} & \hfil20 & \hfil33.21  & \hfil0.964 & \hfil19.3 \\ 	       
			\bottomrule
		\end{tabular}}
	\end{center}
	\label{tab:block_num}
	
\end{table}


\section{Conclusion}
\label{sec:conclusion}
We present a lightweight CNN architecture that we contrast to computationally demanding Transformers that are recently a dominating approach in high-end motion and defocus image deblurring. 
Our core component LaKD block equipped with large kernels leads to a large ERF, resulting in state-of-the-art performance while maintaining simplicity and efficiency. Extensive experiments and ablation studies demonstrate the effectiveness of our method. We additionally propose ERFMeter to quantitatively characterize ERF, which is highly correlated to the network performance. 
However, ERFMeter is an empirical metric of network performance that strongly relies on the ERF paradigm, while ignoring the multitude of other factors. 
Still, we hope it can inspire the community to explore more holistic metrics that could guide efforts toward network performance improvement. In this paper, we only test our network in the image deblurring tasks, while other low-level vision applications like dehazing, deraining, or denoising, \etc could be considered, which we leave as future work. \\
\noindent \textbf{Limitations} \; Our network has a slightly weaker generalization ability than Transformer-based structures as shown in Tab. \ref{tab:motion_comparison_R_J}, despite being partially explained by the compact structure, we still have room for enhancing the generalization ability. Besides, we limit our work to evaluating network performance from the ERF point of view. Future works may further explore other aspects like network representation structure~\cite{raghu2021vision}, loss landscape~\cite{park2022vision}, \etc.

{\small
\bibliographystyle{ieee_fullname}
\bibliography{bibfile}
}

\clearpage

\title{Revisiting Image Deblurring with an Efficient ConvNet \\ - Supplementary Material}
\vspace{-2cm}
\author{}
\date{}

\maketitle

In this supplementary material, we provide more details on the following topics: 
(1) all datasets that we employed in the network training (Sec.\ref{sec:sub:training_details}),
(2) the dilated structure that we consider in the ablation study (Sec.\ref{sec:sub:relevant_dilated}), 
(3) the quantitative comparison on the LFDOF dataset \cite{ruan2021aifnet} (Sec. \ref{sec:sub:result_on_lfdof}),
(4) the effective receptive field (ERF) and its evolution during the training (Sec. \ref{sec:erf_details}), and
(5) additional qualitative comparisons of motion deblurring evaluated on the GoPro \cite{nah2017deep}, HIDE \cite{shen2019human}, and RealBlur \cite{rim2020real} datasets, as well as defocus deblurring evaluated on the DPDD \cite{abuolaim2020defocus}, RealDOF \cite{lee2021iterative}, and CUHK \cite{shi2014discriminative} datasets (Sec. \ref{sec:additional_visual_results}). 

We will make our code and weights publicly available.

\subsection{Training}
\label{sec:sub:training_details}

Here we provide more details on all the training and evaluation datasets for image motion and defocus deblurring tasks we use in this work.\\

\noindent \textbf{Defocus deblurring} We consider four defocus-related datasets. The most popular one is DPDD \cite{abuolaim2020defocus}, which collects blurry-sharp pairs separately with different aperture sizes using a DSLR camera. It has 350 training samples, and 76 testing samples for single-image defocus deblurring, as well as the dual-pixel version for dual-pixel defocus deblurring. RealDOF captures the data pairs in a single shot with a dual-camera setup, offering 50 high-resolution test samples. CUHK is collected for blur detection and provides 706 evaluated low-resolution samples acquired from the Internet. Note that RealDOF and CUHK have only testing samples and, thereby, are good for evaluating the generalization ability. LFDOF is a synthetic defocus blur dataset (11261 training samples and 725 testing samples) generated using a set of light field images, which in our experiment is adopted for the two-stage training strategy as proposed in \cite{ruan2022learning}.

\noindent \textbf{Motion deblurring}  We consider three benchmark datasets -- GoPro \cite{nah2017deep}, HIDE \cite{shen2019human}, and RealBlur \cite{rim2020real}. The GoPro dataset features a synthetic blur that integrates adjacent frames from high-framerate videos to produce motion blur and contains 2013 blurry-sharp training pairs and 1111 testing samples. The HIDE dataset is synthesized following a similar method as the GoPro dataset but emphasizes human-aware deblurring, including large amounts of walking pedestrians, resulting in 6397 training and 2025 testing pairs. Here we follow \cite{wang2022uformer,zamir2022restormer,zamir2021multi} and train our network on the GoPro dataset alone and directly evaluate the HIDE test set for demonstrating the generalization ability. The RealBlur dataset captures real motion blur and sharp images with a dual DSLR camera (Sony A7RM3) setup, which can be obtained simultaneously with different shutter speeds. It offers two subsets sharing the same content, one is output as JPEG images through a camera ISP, and the other is generated as raw images with white balance, demosaicing, denoising, geometric alignment, \etc., resulting in 3758 training samples and 980 testing samples for each set dubbed Real-J and Real-R. In our experiment, we train our network on the GoPro dataset and directly test it on the RealBlur dataset. We also train it on each RealBlur set for extra 450k iterations from the pre-trained weight using GoPro dataset as suggested in \cite{rim2020real} and evaluate their associated testing sets. 

In Tab. S\ref{tab:dataset_details} we summarize the training and testing datasets used in the main manuscript. 

\begin{table}[h!] \scriptsize
	\begin{center}
 	\caption{Summary of training and testing datasets for motion and defocus deblurring tasks.}
        \label{tab:dataset_details}
		\begin{tabular}{@{}p{0.4cm}p{3.8cm}p{1cm}p{2cm}@{}}
		\toprule
			 &\hfil Datasets  &\hfil Training  &\hfil Testing  \\ 
			\midrule
			Motion  &\hfil  GoPro \cite{nah2017deep}  &\hfil 2103 &\hfil 1111   \\ 
			  &\hfil  HIDE \cite{shen2019human}  &\hfil 0 &\hfil 2025       \\ 
			  &\hfil  RealBlur-J\cite{rim2020real} &\hfil 3758 &\hfil 980       \\ 
			  &\hfil  RealBlur-R\cite{rim2020real} &\hfil 3758 &\hfil  980    \\ 
			\midrule
			Defocus &\hfil DPDD \cite{abuolaim2020defocus} &\hfil 350 &\hfil 76     \\ 
                    &\hfil  RealDOF \cite{lee2021iterative} &\hfil 0 &\hfil 50     \\
                    &\hfil  LFDOF \cite{ruan2021aifnet} &\hfil 11261 &\hfil 725     \\ 
                    &\hfil  CUHK \cite{shi2014discriminative} &\hfil 0 &\hfil 704   \\ 
            \midrule
            \midrule
 &\hfil Tab. / Fig.   &\hfil Training  &\hfil Testing  \\ 
\midrule
	Motion &\hfil  Tab. 1, Tab. 2 (upper), Fig. 6,  &\hfil \cite{nah2017deep} &\hfil \cite{nah2017deep} \& \cite{shen2019human} \& \cite{rim2020real}   \\ 
 	 &\hfil  Tab. 7, 9, 10  &\hfil \cite{nah2017deep} &\hfil \cite{nah2017deep}     \\ 
 	 &\hfil  Tab. 2 (lower), Fig. S\ref{fig:motion_realj_0}, S\ref{fig:motion_realr_0} &\hfil \cite{rim2020real} &\hfil \cite{rim2020real}      \\ 
  	 &\hfil Fig. S\ref{fig:motion_gopro_0}, S\ref{fig:motion_hide_0} &\hfil \cite{nah2017deep} &\hfil \cite{nah2017deep} \cite{shen2019human}    \\ 
\midrule

          	Defocus &\hfil  Tab. 3, 4, 6 -- 10, S\ref{tab:layer_ablate}, Fig. 5, S\ref{fig:dpdd_0}, S\ref{fig:realdof_0} &\hfil\cite{abuolaim2020defocus} &\hfil\cite{abuolaim2020defocus} \& \cite{lee2021iterative}    \\ 

          
             &\hfil  Tab. 5, Fig. 7, S\ref{fig:dpdd_two_stage_0}, S\ref{fig:realdof_two_stage_0},S\ref{fig:0_cuhk}, S\ref{fig:1_cuhk} &\hfil\cite{abuolaim2020defocus} \&\hfil\cite{ruan2021aifnet}   & \hfil \cite{abuolaim2020defocus}, \cite{lee2021iterative},\cite{shi2014discriminative}\\    
             &\hfil  Tab. S\ref{tab:lfdof_result} &\hfil \cite{ruan2021aifnet}   & \hfil \cite{ruan2021aifnet}\\ 
			\bottomrule
		\end{tabular}
	\end{center}
	
\end{table}

\subsection{Ablation of network structure}
\label{sec:sub:relevant_dilated}

Here we ablate our network with respect to its structure and layer number. \\

\begin{figure}[htb]
    \setlength{\abovecaptionskip}{0.1cm}
    \setlength{\belowcaptionskip}{0.1cm}
	\centering
	\includegraphics[width=0.9\linewidth]{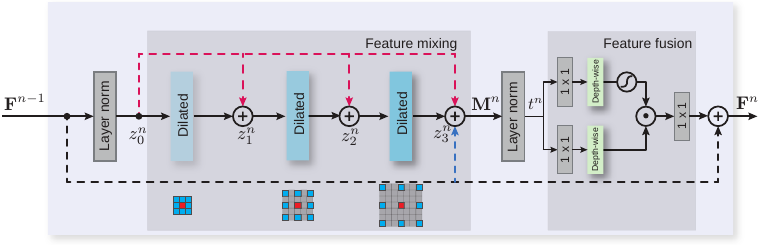}
	\caption{The structure of dilated convolution layers that can be compared with the corresponding structure in the LaKD block (refer to Fig. 2 in the main manuscript).}
	\label{fig:dilated_abalte}
\end{figure}

\noindent \textbf{Version with dilated convolution layers} 
As opposed to the LaKD block described in the main paper, here, we present its alternative version with dilated convolution layers (Fig. S\ref{fig:dilated_abalte}). 
Note that both versions aim to expand the effective receptive field. The dilated version adopts the same structure except for the feature mixing module that consists of three dilated convolution layers with increasing dilation rates. The results in Tab. 7 (refer to the main manuscript) further indicate the superiority of LaKD block. \\

\noindent \textbf{Layer number}{ } We ablate the layer number required in the feature mixing module, specifically on depthwise and pointwise convolution. For example, the label ``one'' in Tab. S\ref{tab:layer_ablate} denotes that the feature mixing module has one depthwise and one pointwise convolution layer. Table S\ref{tab:layer_ablate} shows that feature mixing equipped with two sequential depthwise and pointwise layers can reach the best performance.

		\begin{table}[htb] 
		\caption{Layer number ablation in the feature mixing module. }
        \vspace{-0.5cm}
		\label{tab:layer_ablate}
		\begin{center}
		\resizebox{0.85\linewidth}{!}{%
			\begin{tabular}{@{}p{2.5cm}p{1.8cm}p{1.8cm}p{1.8cm}p{1cm}p{1cm}@{}}
				
				\toprule
				\multirow{2}{*}[-0.45em]{\hfil Number}& \multicolumn{3}{c}{DPDD} &  \multirow{2}{*}[-0.0em]{\makecell{Params.\\(M)}} & \multirow{2}{*}[-0.0em]{\makecell{MACs\\(G)}}\\ 
				\cmidrule(rl){2-4}                      
				& \hfil PSNR$\uparrow$  &\hfil SSIM$\uparrow$  &\hfil LPIPS$\downarrow$  \\ 
				\midrule
				
				\hfil one  &  \hfil 26.10  & \hfil 0.808 & \hfil 0.155   &  \hfil 15.4 & \hfil 1004\\   
				\hfil two (ours)& \hfil  26.15 & \hfil 0.810 & \hfil 0.155 & \hfil 17.7 & \hfil 1208    \\ 
    	        \hfil three & \hfil 26.09  & \hfil 0.806 & \hfil 0.156  & \hfil 20.1  &\hfil 1413\\ 
				\bottomrule
			\end{tabular}}
		\end{center}

	\end{table}

\subsection{Performance on the LFDOF dataset}
\label{sec:sub:result_on_lfdof}
Following \cite{ruan2022learning}, we additionally investigate our network performance on the LFDOF dataset.  AIFNet \cite{ruan2021aifnet} has two subnets, sharing a similar spirit to the conventional methods that explicitly estimate the defocus map and then perform non-blind deconvolution. DRBNet \cite{ruan2022learning} adopts an end-to-end solution and resorts to per-pixel kernel estimation to account for the spatially-varying blur. Our method is also an end-to-end solution that employs the LaKD block with a large effective receptive field, which leads to much better performance, as shown in Tab. S\ref{tab:lfdof_result}.

		\begin{table}[htb] 
		\caption{Quantitative comparison between AIFNet \cite{ruan2021aifnet}, DRBNet \cite{ruan2022learning}, and our network evaluated on 725 images from the LFDOF test set. }
        \vspace{-0.5cm}
		\label{tab:lfdof_result}
		\begin{center}
		\resizebox{0.7\linewidth}{!}{%
			\begin{tabular}{@{}p{2.5cm}p{1.8cm}p{1.8cm}p{1.8cm}@{}}
				
				\toprule
				\multirow{2}{*}[-0.45em]{ Method}& \multicolumn{3}{c}{LFDOF}  \\ 
				\cmidrule(rl){2-4}                      
				
				& PSNR$\uparrow$  & SSIM$\uparrow$  & LPIPS$\downarrow$  \\ 
				\midrule
				
				AIFNet \cite{ruan2021aifnet} &  29.69  & 0.880 & 0.151      \\   
				DRBNet \cite{ruan2022learning}& 30.40 & 0.891 & 0.145     \\ 
    	        Ours & \textbf{31.87}  & \textbf{0.912} & \textbf{0.115}      \\ 
				
				\bottomrule
			\end{tabular}}
		\end{center}

	\end{table}

\section{ERF fitting details}
\label{sec:erf_details}

\begin{figure*}[htb] 
	\centering
	\includegraphics[width=\linewidth]{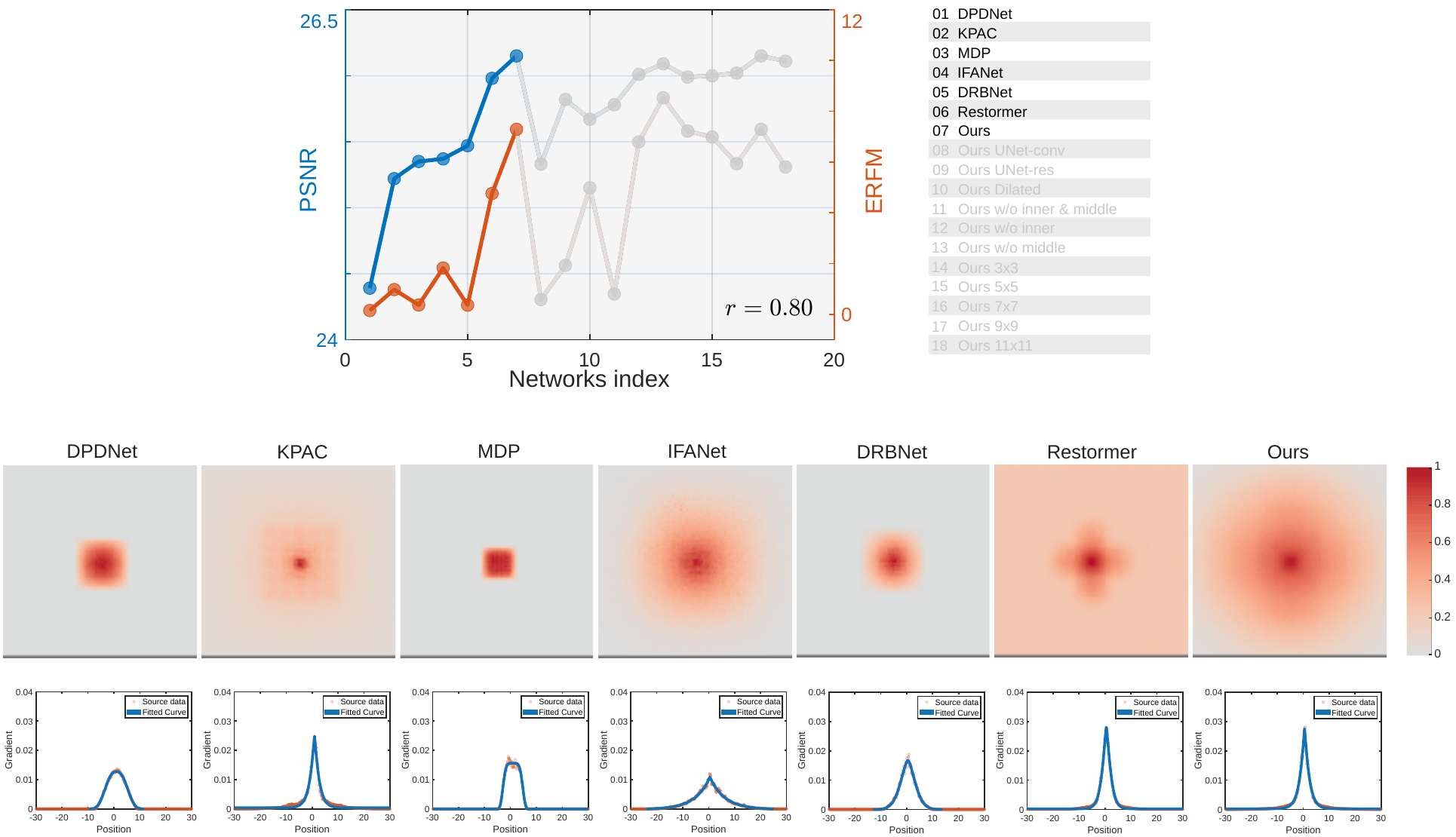}
	\caption{Demonstration of ERF patterns and the fitted GND-PDF
curves:  Here, different defocus deblurring networks are considered. Note that, the ERF patterns are in log scale for better visualization, while the GND-PDF curves are in linear scale.}
	\label{fig:erf_defocus}
\end{figure*}

\begin{figure*}[htb] 
	\centering
	\includegraphics[width=\linewidth]{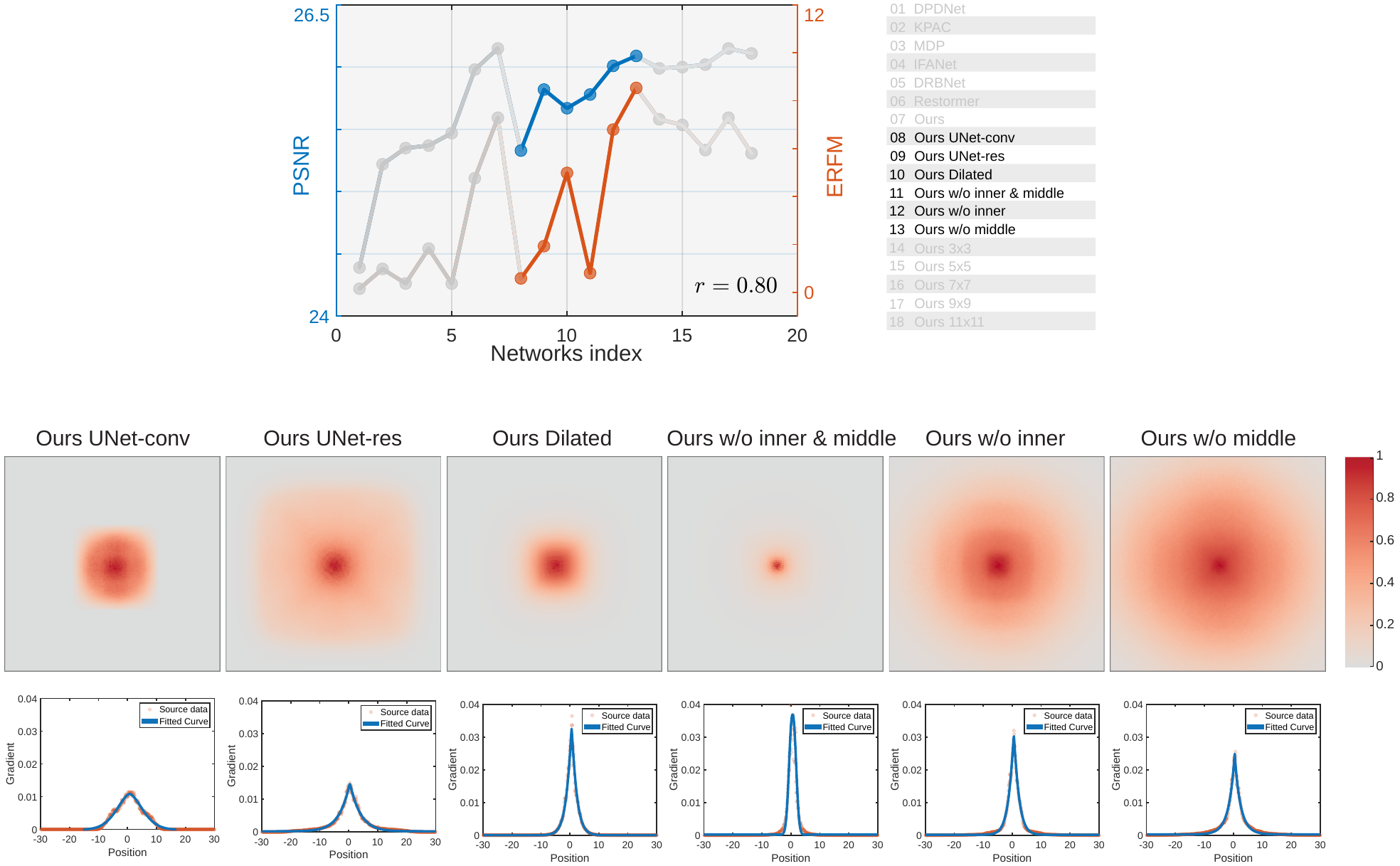}
	\caption{Demonstration of ERF patterns and the fitted GND-PDF curves: Here, different variants of our network are considered.}
	\label{fig:erf_ablation}
\end{figure*}

In this section, we provide more details of our ERF visualization, GND-PDF fitting, and ERFMeter. We visualize all the ERFs and GND-PDF fittings of defocus deblurring networks \cite{abuolaim2020defocus,son2021single,abuolaim2022improving,ruan2022learning,zamir2022restormer} in Fig.~S\ref{fig:erf_defocus}, variants of our networks that used for ablation study in Fig.~S\ref{fig:erf_ablation}, and our network with different kernel sizes in Fig.~S\ref{fig:erf_kernel}, correspondingly, all the parameters of GND-PDF fitting could be find in Tab.~S\ref{tab:param_defocus}, Tab.~S\ref{tab:param_ablation}, and Tab.~S\ref{tab:param_kernel}. Besides, we also select several representative networks for motion deblurring \cite{kupyn2018deblurgan,gao2019dynamic,cho2021rethinking,zamir2022restormer} and visualize their ERFs in Fig.~S\ref{fig:erf_motion}. Note that the network structures are highly diverse, especially for networks on motion deblurring \eg multi-patch \cite{zhang2019deep}, multi-scale \cite{nah2017deep}, recurrent scheme \cite{park2020multi}. In this paper, we intend to reduce the diversity and only select networks with overall U-Net architecture so that the visualized ERFs are comparable.

We investigate the ERF \cite{luo2016understanding} on the feature extracted from their bottleneck layer (the layer right before the first up-sampling in the decoder), which could potentially reveal the largest ERF they can achieve. The layer names corresponding to the bottleneck layer in each method are list in Tab.~S\ref{tab:param_defocus} to Tab.~S\ref{tab:param_motion}. The ERFs of networks for defocus deblurring are averaged from 912 image patches in size $512\times512$, which are augmented from 76 testing images in DPDD dataset \cite{abuolaim2020defocus} to further eliminate the dependence on input content, while ERFs of networks for motion deblurring are averaged from 1111 image patches in size $512\times512$ from GoPro dataset \cite{nah2017deep}. When doing GND-PDF curve fitting, the x-axis is empirically scaled from $[1, 512]$ to $[-30, 30]$ for higher fitting accuracy. We show the goodness of fitting $R^2$ for each network in Tab.~S\ref{tab:param_defocus} to Tab.~S\ref{tab:param_motion}.

\noindent \textbf{ERF evolution during training} We additionally demonstrate the ERF evolution during training for the motion and defocus deblurring tasks, as shown in Fig.  S\ref{fig:erf_evo_training}. The ERF expands progressively with the training iterations and becomes much larger than at the initial stages. This observation is aligned with \cite{luo2016understanding}.

\section{Additional qualitative results}
\label{sec:additional_visual_results}
 In this section we show more qualitative results on motion and defocus deblurring. Note that we mainly compare our method with Restormer \cite{zamir2022restormer} as it achieves state-of-art performance. \\

\noindent \textbf{Motion deblurring}{}  We include additional visual results that are obtained using image samples from the GoPro (Fig. S\ref{fig:motion_gopro_0}), HIDE (Fig. S\ref{fig:motion_hide_0}), Real-J (Fig. S\ref{fig:motion_realj_0}), and Real-R (Fig. S\ref{fig:motion_realr_0}) datasets. Those results complement Tabs. 1 and 2 in the main manuscript. \\
\noindent \textbf{Defocus deblurring}{} We include visual results for single-image defocus deblurring for image samples from the DPDD (Fig. S\ref{fig:dpdd_0}) and RealDOF (Fig. S\ref{fig:realdof_0}) datasets. Those results complement Tab. 3.  We also provide visual results for dual-pixel defocus deblurring for image samples from the DPDD (Fig. S\ref{fig:0_dual_dpdd}) dataset, which complement Tab. 4. We further compare our method with DRBNet \cite{ruan2022learning} adopting the two-stage training strategy proposed in \cite{ruan2022learning}, and we evaluate both methods using image samples from the DPDD (Fig. S\ref{fig:dpdd_two_stage_0}), RealDOF (Fig. S\ref{fig:realdof_two_stage_0}), and CUHK (Fig. S\ref{fig:0_cuhk}, S\ref{fig:1_cuhk}) datasets, which  complement Tab. 5.

\begin{table*}[htb] \scriptsize
    \caption{ERF fitting parameters and statistics for defocus deblurring networks.}  
	\begin{center}
        \resizebox{\textwidth}{!}{
		\begin{tabular}{@{}>{\columncolor{white}[0pt][\tabcolsep]}p{2cm}p{2cm}p{0.8cm}p{0.8cm}p{0.8cm}p{0.8cm}p{0.8cm}p{0.8cm}p{0.8cm}>{\columncolor{white}[\tabcolsep][0pt]}p{0.8cm}@{}}
	\toprule
                
        Method & Layer Name & \hfil $\sigma$ & \hfil$\beta$ & \hfil$\mu$ & \hfil$c_1$ & \hfil$c_2$ (e-5) & \hfil$R^2$ & \hfil PSNR& \hfil ERFM\\
		\midrule  
         DPDNet \cite{abuolaim2020defocus} & \textsf{conv5\_2} & \hfil 5.2054 & \hfil 2.6054 &\hfil 0.9550   &\hfil  0.1188 &\hfil -2.61 &\hfil  0.9978 &\hfil 24.39 &\hfil 0.1469 \\ 
         KPAC \cite{son2021single}     & \textsf{conv4\_4} &\hfil 1.8945 & \hfil 1.0838 &\hfil  0.8993   &\hfil  0.0923 &\hfil  41.47   &\hfil  0.9774  &\hfil 25.22 &\hfil 0.9737 \\ 
         MDP \cite{abuolaim2022improving}  & \textsf{conv15} &\hfil 4.0575 & \hfil 4.9008 &\hfil  0.9371   &\hfil  0.1174 &\hfil -0.32   &\hfil0.9956  &\hfil 25.35 &\hfil 0.3658 \\
        IFANet\cite{lee2021iterative}     & \textsf{conv\_res} &\hfil 5.6632 & \hfil 0.9690 &\hfil 0.4738   &\hfil  0.1291 &\hfil  -18.31   &\hfil  0.9885  &\hfil 25.37 &\hfil 1.8221\\ 
        DRBNet \cite{ruan2022learning}       & \textsf{conv4\_4} &\hfil 3.8936 &\hfil  1.5960 &\hfil 0.4254 &\hfil  0.1183 &\hfil  -1.79    &\hfil   0.9942 &\hfil 25.47 &\hfil 0.3615 \\
        Restormer \cite{zamir2022restormer}  & \textsf{latent} &\hfil  1.9964 & \hfil  1.2687 &\hfil   0.5252   &\hfil  0.1057 &\hfil  19.14    &\hfil 0.9953 &\hfil 25.98 &\hfil 4.7581\\ 
        Ours  & \textsf{bt\_neck} &\hfil  1.9138 & \hfil  1.0812 &\hfil   0.5105   &\hfil  0.1044 &\hfil  21.27    &\hfil 0.9914 &\hfil 26.15 &\hfil 7.2870\\

         \bottomrule
		\end{tabular}}
	\end{center}
        \label{tab:param_defocus}

\end{table*}

\clearpage

\begin{table*}[htb] \scriptsize
    \caption{ERF fitting parameters and statistics for variants of our network.}
	\begin{center}
        \resizebox{\textwidth}{!}{
		\begin{tabular}{@{}>{\columncolor{white}[0pt][\tabcolsep]}p{2cm}p{2cm}p{0.8cm}p{0.8cm}p{0.8cm}p{0.8cm}p{0.8cm}p{0.8cm}p{0.8cm}>{\columncolor{white}[\tabcolsep][0pt]}p{0.8cm}@{}}
	\toprule
                
        Method & Layer Name & \hfil $\sigma$ & \hfil$\beta$ & \hfil$\mu$ & \hfil$c_1$ & \hfil$c_2$ (e-5) & \hfil$R^2$ & \hfil PSNR& \hfil ERFM\\
		\midrule  

	UNet-conv  & \textsf{bt\_neck} &\hfil 6.2515 &\hfil  1.6787 &\hfil  0.7357  &\hfil 0.1224 &\hfil -8.73    &\hfil   0.9841 &\hfil 25.33 &\hfil 0.5774\\

        UNet-res  & \textsf{bt\_neck} &\hfil 3.6274 &\hfil 1.0466 &\hfil  0.4065  &\hfil 0.1048 &\hfil 20.60  &\hfil  0.9873 &\hfil 25.82 &\hfil 1.9244 \\
         Ours-dilated  & \textsf{bt\_neck} &\hfil 1.8397 &\hfil  1.1123 &\hfil  0.5869  &\hfil 0.1170 &\hfil  0.30  &\hfil  0.9812  &\hfil 25.67 & \hfil 4.9839 \\
          Ours w/o both  & \textsf{bt\_neck} &\hfil 1.6120 &\hfil  2.1925 &\hfil 0.5965  &\hfil 0.1067 &\hfil  17.43  &\hfil  0.9047 &\hfil 25.78 & \hfil 0.7989 \\
         Ours w/o inner   &  \textsf{bt\_neck} &\hfil 1.8337 &\hfil  1.1069 &\hfil 0.5574  &\hfil 0.1089 &\hfil  13.79  &\hfil   0.9916 &\hfil 26.01 &\hfil 6.7925 \\
       Ours w/o middle  & \textsf{bt\_neck} &\hfil 1.9742 &\hfil  0.9474 &\hfil 0.4418 &\hfil 0.1025 &\hfil  24.41  &\hfil   0.9845 &\hfil 26.09 &\hfil 8.5311 \\
        
         \bottomrule
		\end{tabular}}
	\end{center}
    \label{tab:param_ablation}
	
\end{table*}

\clearpage

\begin{figure*}[htb] 
	\centering
	\includegraphics[width=\linewidth]{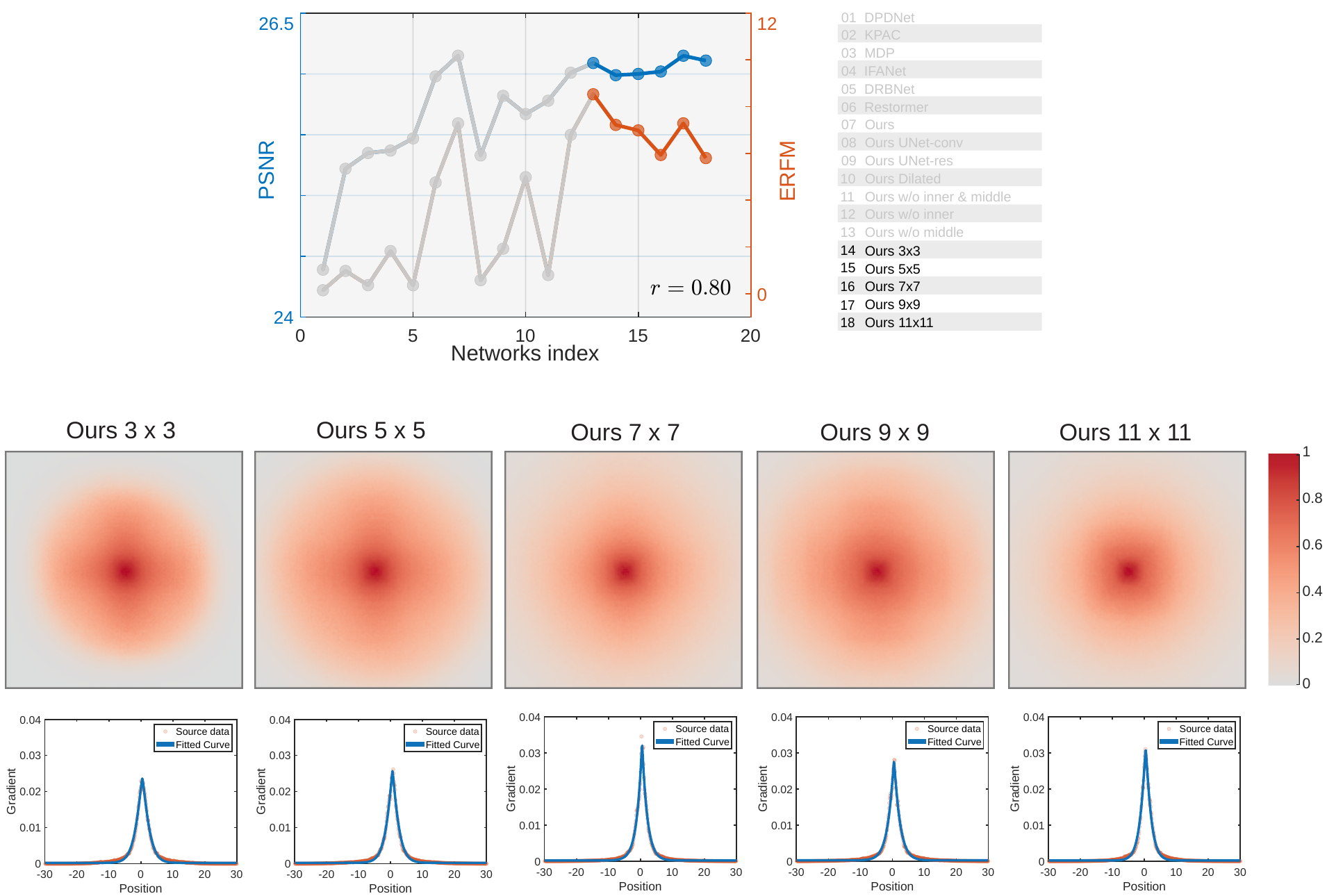}
	\caption{Demonstration of ERF patterns and the fitted GND-PDF
curves: Here, we consider our network with different kernel sizes.}
	\label{fig:erf_kernel}
\end{figure*}

\begin{table*}[htb] \scriptsize
    \caption{ERF fitting parameters and statistics for our network with different kernel sizes.}  
	\begin{center}
        \resizebox{\textwidth}{!}{
		\begin{tabular}{@{}>{\columncolor{white}[0pt][\tabcolsep]}p{2cm}p{2cm}p{0.8cm}p{0.8cm}p{0.8cm}p{0.8cm}p{0.8cm}p{0.8cm}p{0.8cm}>{\columncolor{white}[\tabcolsep][0pt]}p{0.8cm}@{}}
	\toprule
                
        Method & Layer Name & \hfil $\sigma$ & \hfil$\beta$ & \hfil$\mu$ & \hfil$c_1$ & \hfil$c_2$ (e-5) & \hfil$R^2$ & \hfil PSNR& \hfil ERFM\\
        \midrule
        ours ($3 \times 3$)  & \textsf{bt\_neck} &\hfil 2.3390 &\hfil  1.2508 &\hfil  0.4832  &\hfil 0.1032 &\hfil  23.31   &\hfil 0.9928 &\hfil 25.99 &\hfil 7.2201 \\
        ours ($5 \times 5$)  & \textsf{bt\_neck} &\hfil 2.0966 &\hfil  1.1722 &\hfil 0.5941 &\hfil 0.1024 &\hfil  24.59 &\hfil  0.9912 &\hfil 26.00 &\hfil  6.9847 \\
        ours ($7 \times 7$) & \textsf{bt\_neck} &\hfil 1.6297 &\hfil 1.0267 &\hfil 0.5398 &\hfil 0.1071 &\hfil 16.86  &\hfil  0.9863 &\hfil 26.02 &\hfil 5.9340 \\
        ours ($9 \times 9$) & \textsf{bt\_neck} &\hfil 1.9138 &\hfil 1.0812 &\hfil 0.5105 &\hfil 0.1044 &\hfil 21.27 &\hfil  0.9914 &\hfil 26.15 &\hfil 7.2870 \\
        
        ours ($11 \times 11$) & \textsf{bt\_neck} &\hfil 1.7744 &\hfil 1.1446 &\hfil 0.4124 &\hfil 0.1067 &\hfil 17.42 &\hfil  0.9924 &\hfil 26.11 &\hfil 5.8027 \\
        
         \bottomrule
		\end{tabular}}
	\end{center}

	\label{tab:param_kernel}
\end{table*}

\clearpage

\begin{figure*}[htb] 
	\centering
	\includegraphics[width=\linewidth]{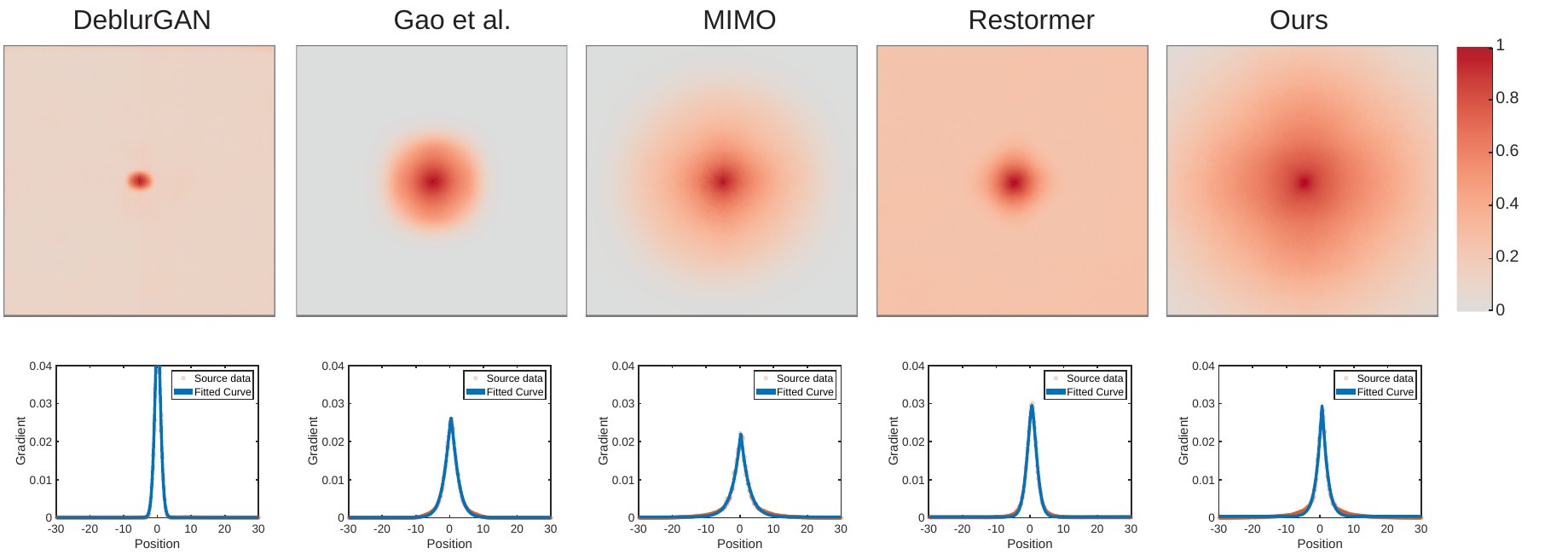}
	\caption{Demonstration of ERF patterns and the fitted GND-PDF
curves: Here, different motion deblurring networks are considered.}
	\label{fig:erf_motion}
\end{figure*}

\begin{table*}[htb] \scriptsize
    \caption{ERF fitting parameters and statistics for motion deblurring networks.}   
	\begin{center}
        \resizebox{\textwidth}{!}{
		\begin{tabular}{@{}>{\columncolor{white}[0pt][\tabcolsep]}p{2cm}p{2cm}p{0.8cm}p{0.8cm}p{0.8cm}p{0.8cm}p{0.8cm}p{0.8cm}p{0.8cm}>{\columncolor{white}[\tabcolsep][0pt]}p{0.8cm}@{}}
	\toprule
                
        Method & Layer Name & \hfil $\sigma$ & \hfil$\beta$ & \hfil$\mu$ & \hfil$c_1$ & \hfil$c_2$ (e-5) & \hfil$R^2$ & \hfil PSNR& \hfil ERFM\\
        \midrule
         DeblurGAN \cite{kupyn2018deblurgan}  & \textsf{ResnetBlock\_18} & \hfil 1.1834 & \hfil 1.6237 &\hfil 0.0865   &\hfil  0.1145   &\hfil 4.44  &\hfil 0.9563 & \hfil 28.70 &\hfil 2.4235\\ 
         Gao \etal \cite{gao2019dynamic}  & \textsf{level3\_deconv3\_1} & \hfil 2.2912 & \hfil 1.2337 &\hfil 0.4046   &\hfil  0.1142 &\hfil  5.00&\hfil 0.9967  & \hfil 30.90 &\hfil 5.3920 \\ 
         MIMO-UNet+ \cite{cho2021rethinking}   & \textsf{DB3} &\hfil 2.4514 & \hfil 1.0489 &\hfil 0.2753  &\hfil  0.1077  &\hfil 15.78 &\hfil 0.9902 & \hfil 32.45 &\hfil 5.3514\\ 
         Restormer \cite{zamir2022restormer}   & \textsf{latent} &\hfil 1.9640 & \hfil 1.5154 &\hfil 0.5592  &\hfil  0.1054  &\hfil 19.63 &\hfil 0.9952 &\hfil 32.92 &\hfil 3.8813\\ 
        Ours        & \textsf{bt\_neck} &\hfil 1.7278 &\hfil  1.1311 &\hfil  0.5805  &\hfil  0.0977 &\hfil  32.52  &\hfil   0.9863 &\hfil 33.35 &\hfil 6.3704\\
         \bottomrule
		\end{tabular}}
	\end{center}

	\label{tab:param_motion}
\end{table*}
\clearpage

\begin{figure*}[htb] 
	\centering
	\includegraphics[width=0.9\linewidth]{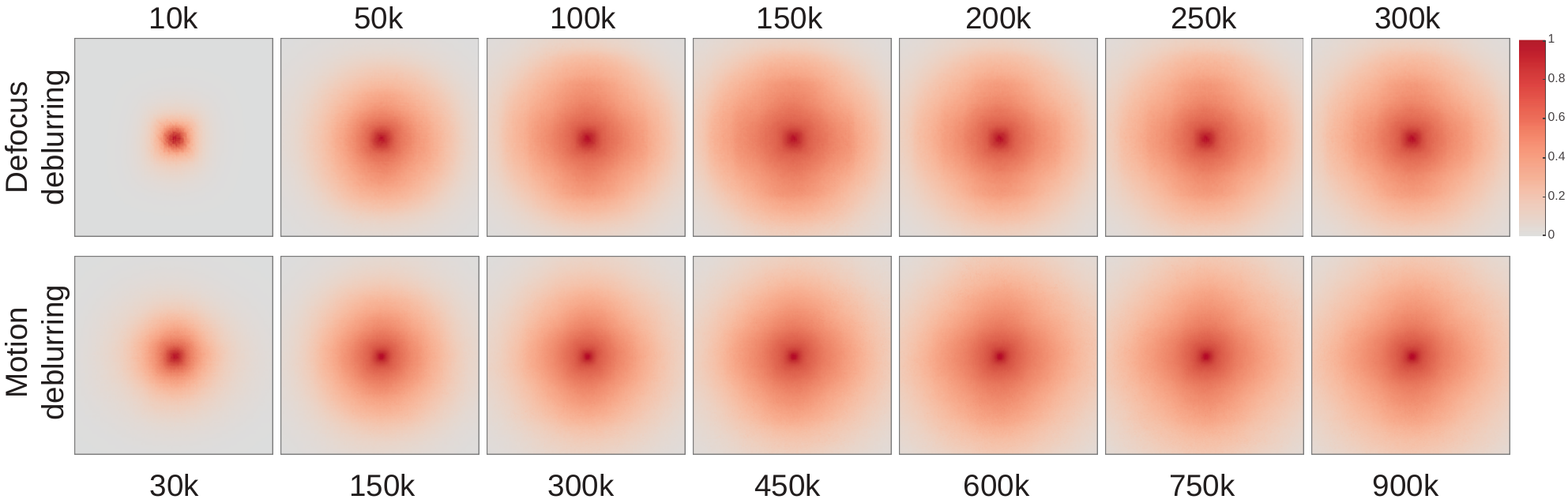}
	\caption{ERF evolution for increasing number of training cycles (from left to right) on defocus deblurring (up) and motion deblurring (down).}
	\label{fig:erf_evo_training}
\end{figure*}

\begin{figure*}[htb] 
	\centering
	\includegraphics[width=0.9\linewidth]{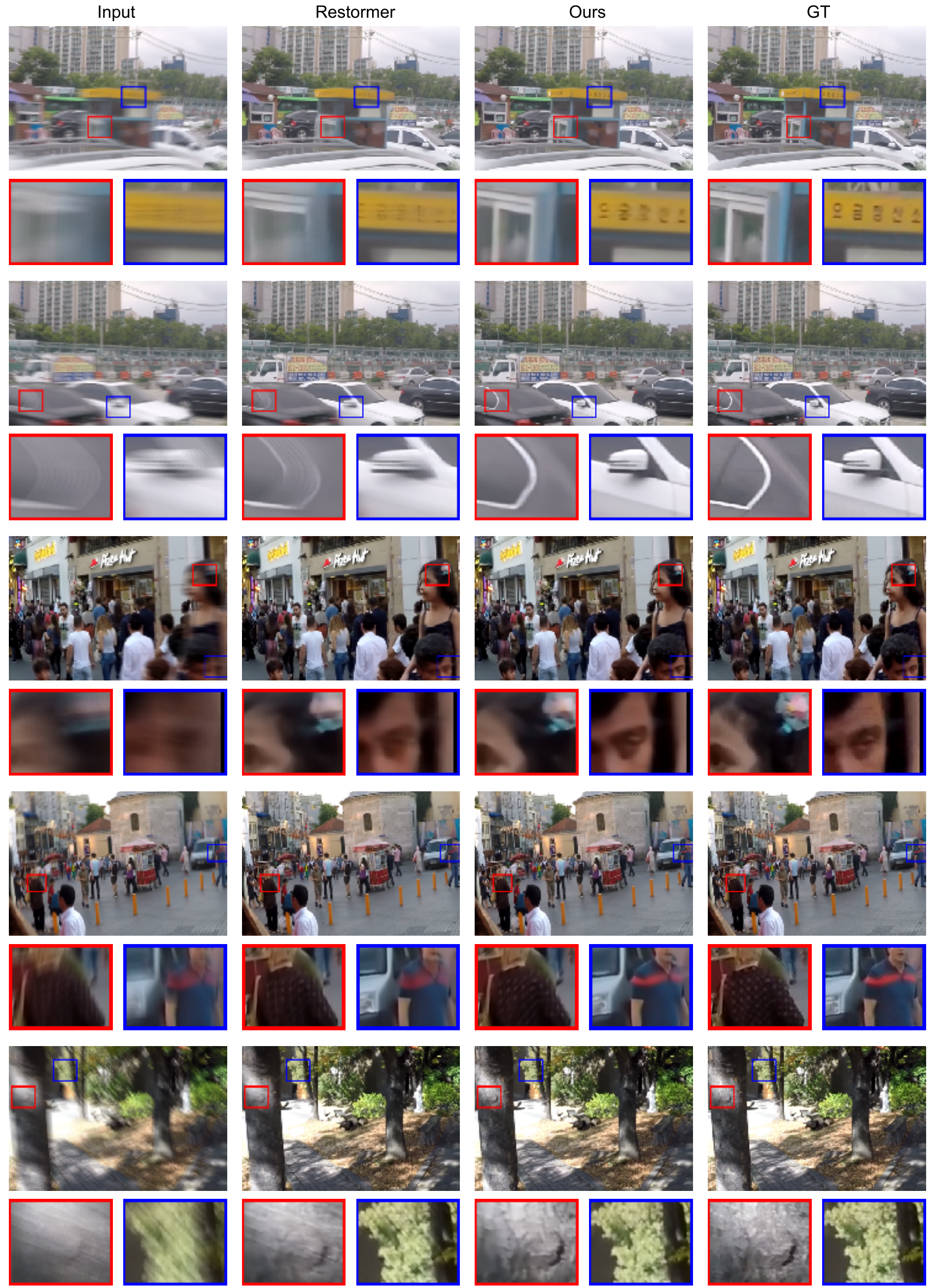}
	\caption{Qualitative comparison between Restormer \cite{zamir2022restormer} and our method evaluated on GoPro dataset.}
	\label{fig:motion_gopro_0}
\end{figure*}

\begin{figure*}[htb] 
	\centering
	\includegraphics[width=0.9\linewidth]{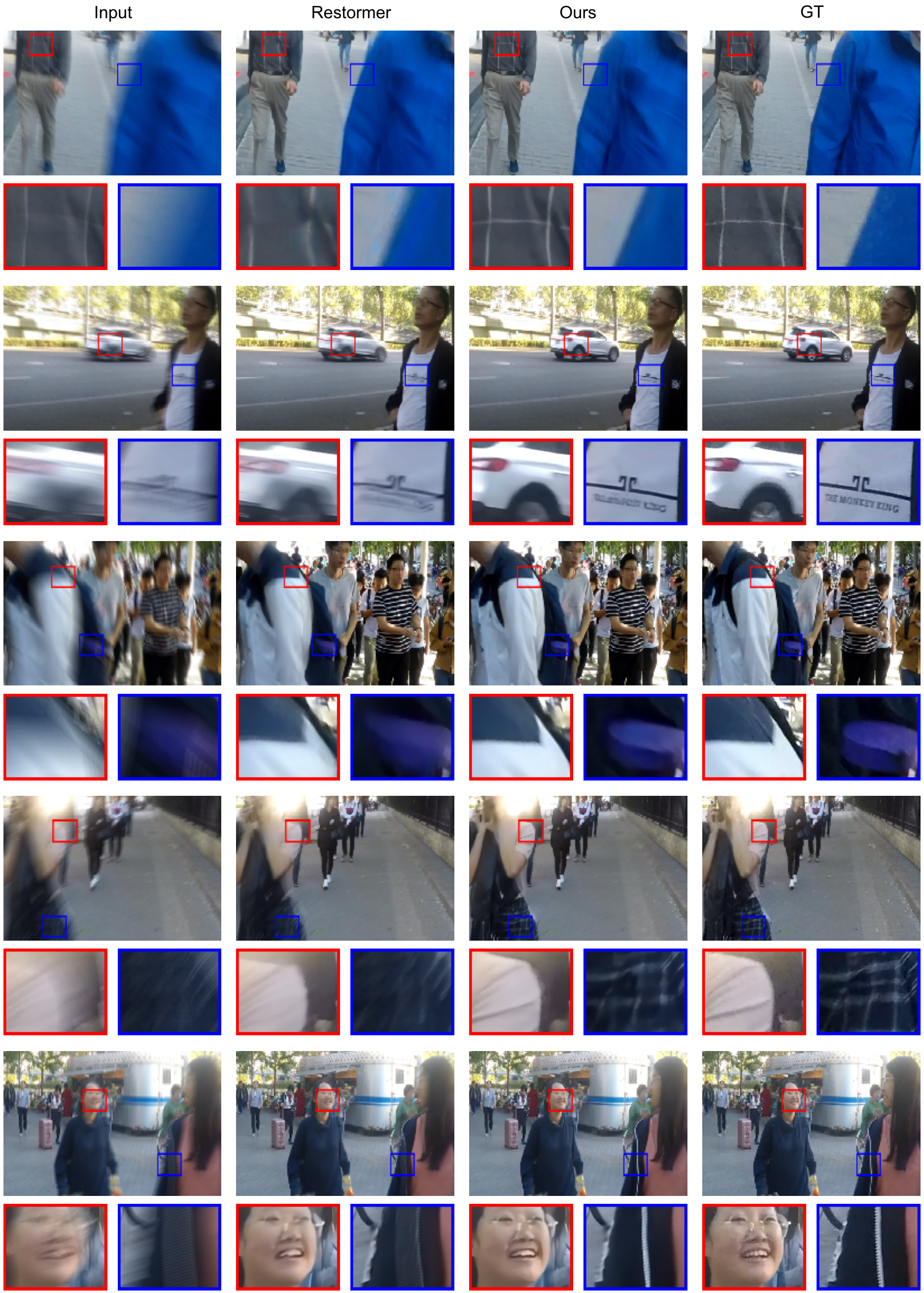}
	\caption{Qualitative comparison between Restormer \cite{zamir2022restormer} and our method evaluated on HIDE dataset.}
	\label{fig:motion_hide_0}
\end{figure*}

\begin{figure*}[htb] 
	\centering
	\includegraphics[width=0.8\linewidth]{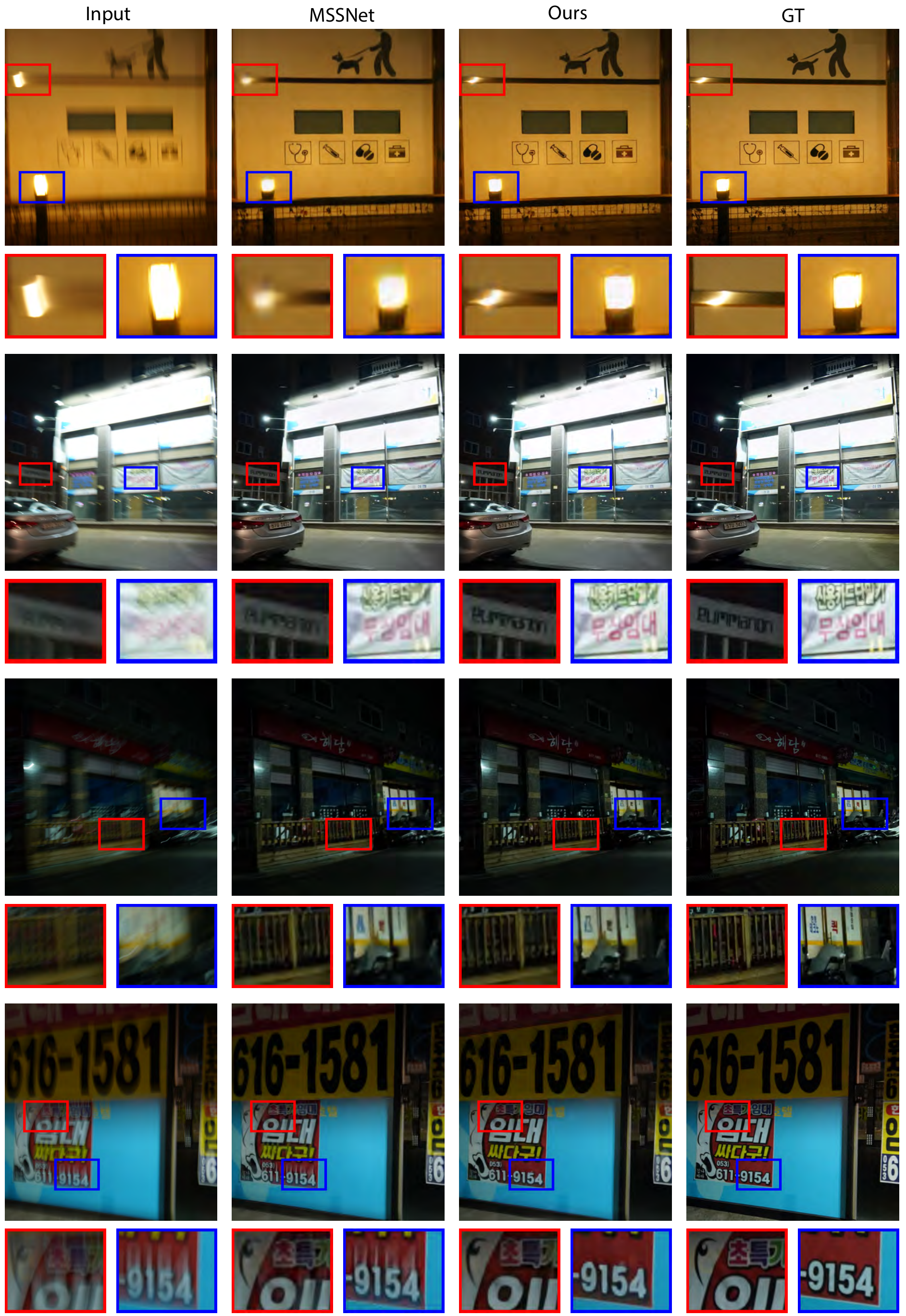}
	\caption{Qualitative comparison between MSSNet \cite{kim2022mssnet} and our method evaluated on RealBlur-J dataset.}
	\label{fig:motion_realj_0}
\end{figure*}

\begin{figure*}[htb] 
	\centering
	\includegraphics[width=0.8\linewidth]{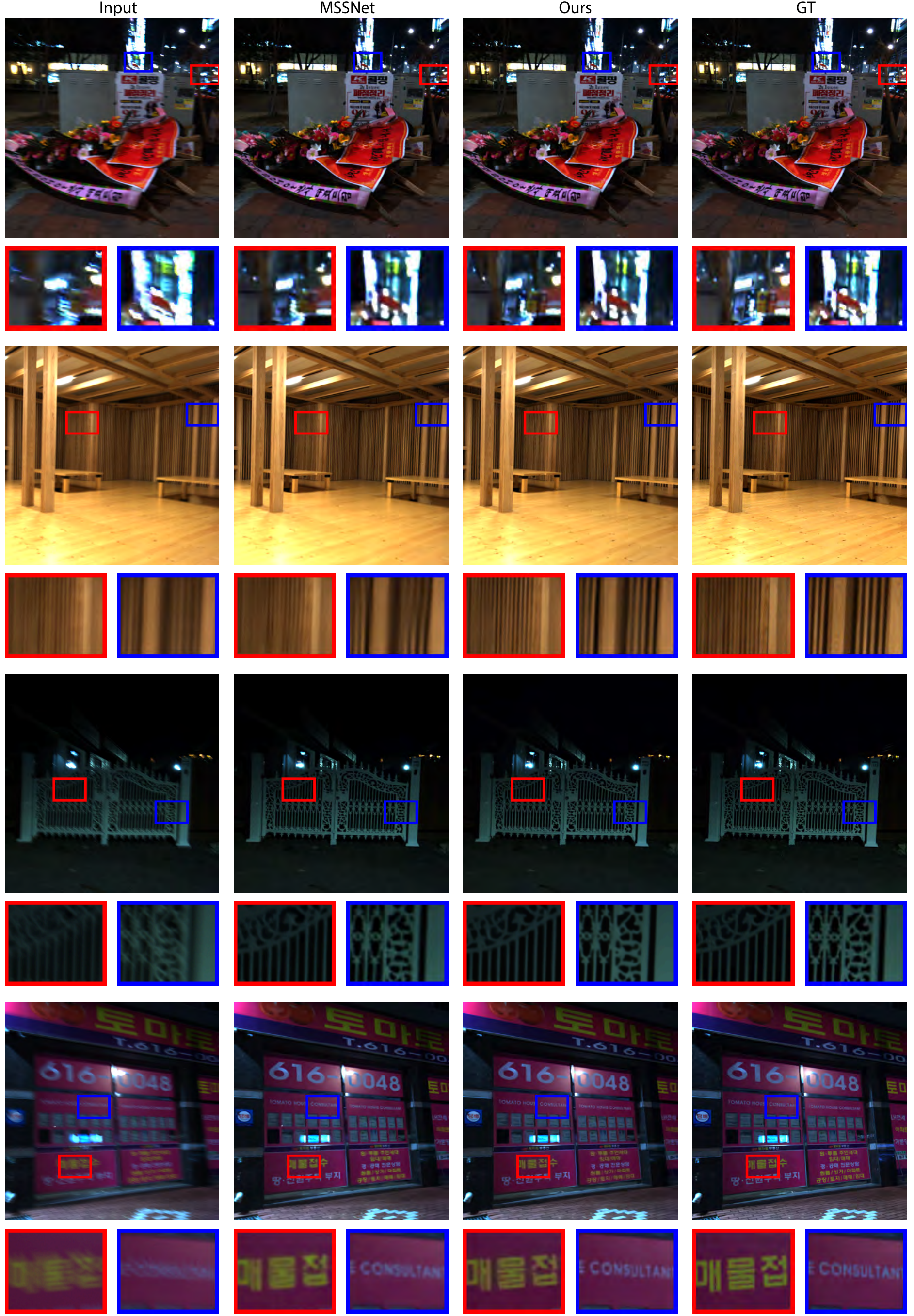}
	\caption{Qualitative comparison between MSSNet \cite{kim2022mssnet} and our method evaluated on RealBlur-R dataset.}
	\label{fig:motion_realr_0}
\end{figure*}

\begin{figure*}[htb] 
	\centering
	\includegraphics[width=0.9\linewidth]{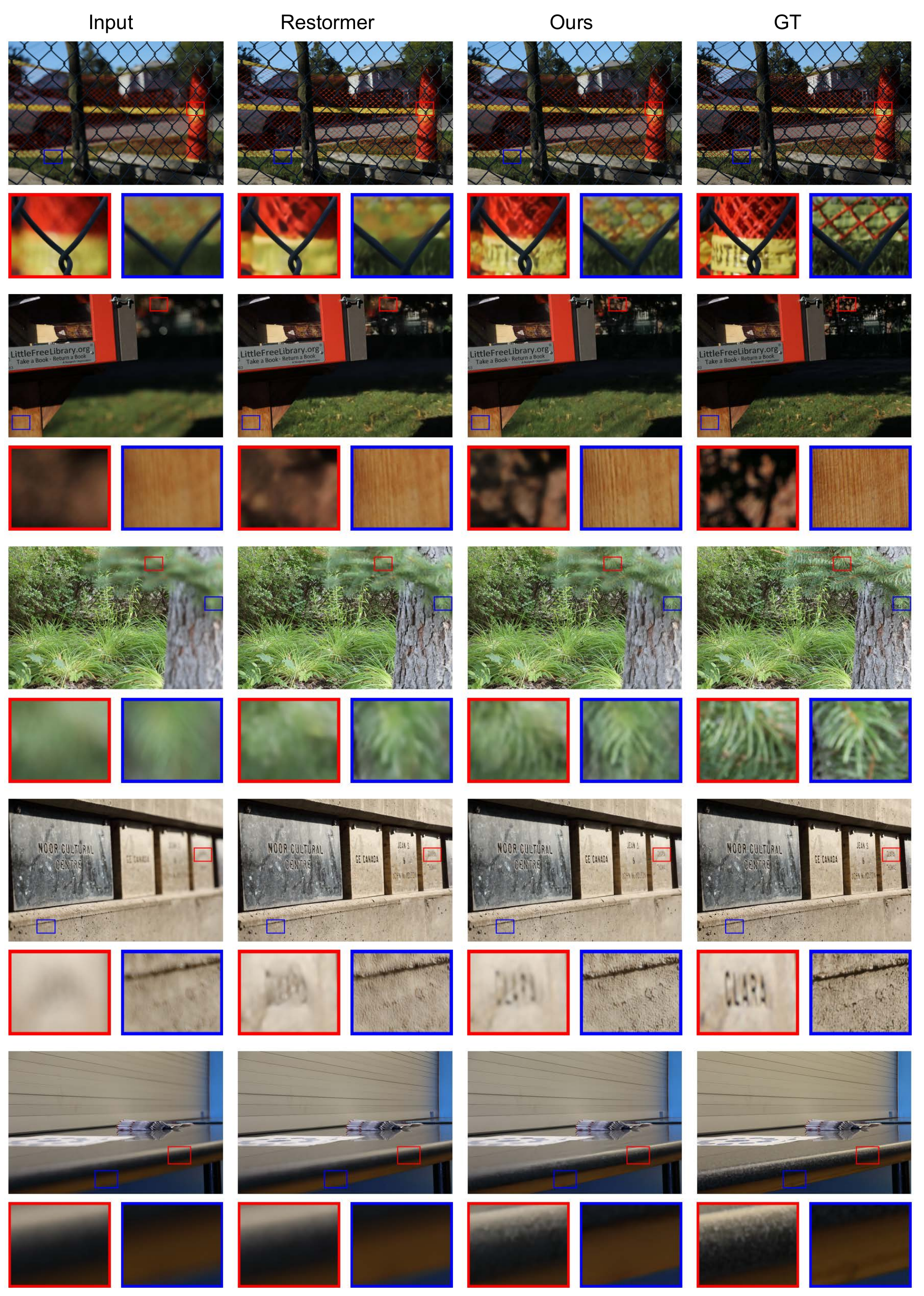}
	\caption{Qualitative comparison between Restormer \cite{zamir2022restormer} and our method evaluated on DPDD dataset.}
	\label{fig:dpdd_0}
\end{figure*}

\begin{figure*}[htb] 
	\centering
	\includegraphics[width=0.9\linewidth]{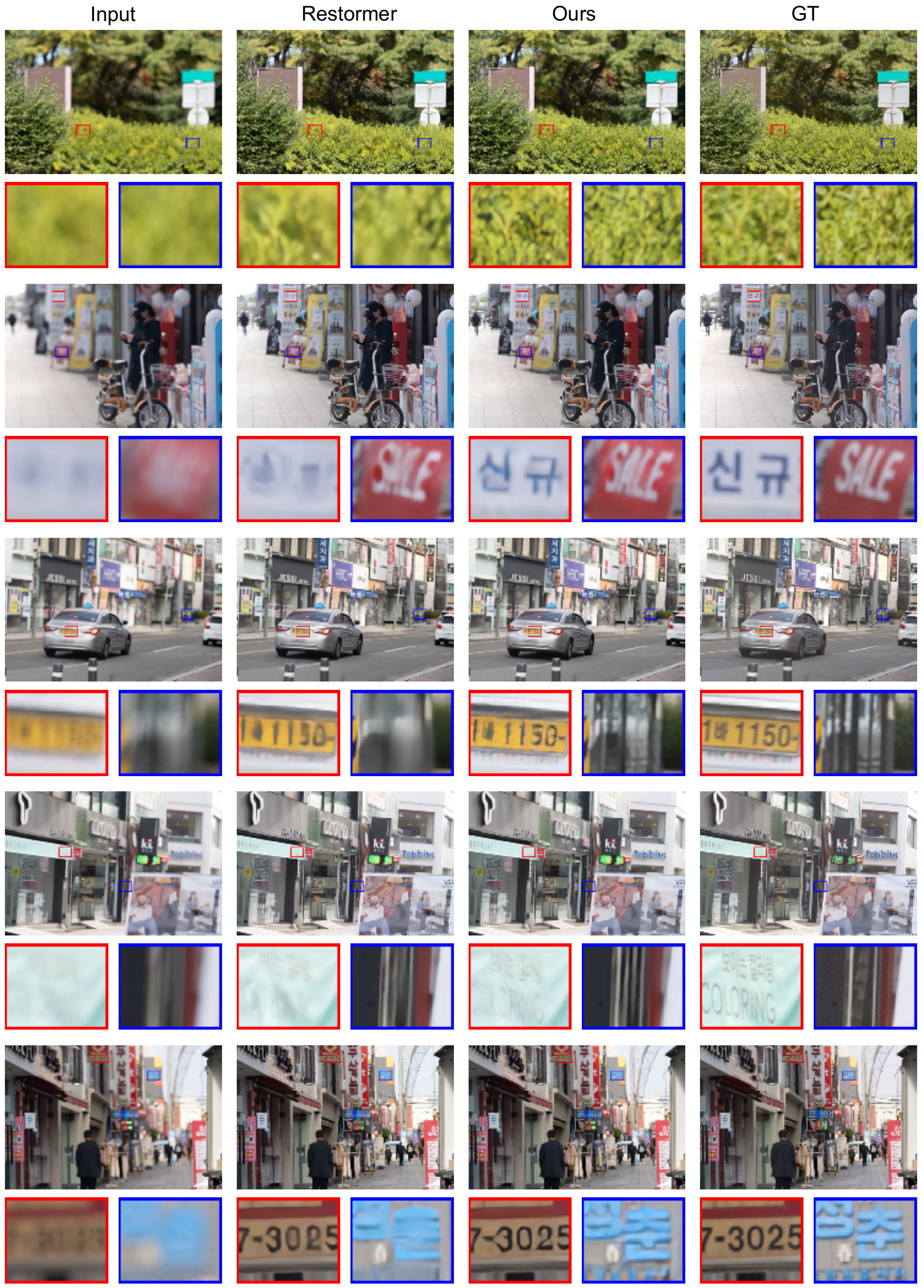}
	\caption{Qualitative comparison between Restormer \cite{zamir2022restormer} and our method evaluated on RealDOF dataset.}
	\label{fig:realdof_0}
\end{figure*}

\begin{figure*}[htb] 
	\centering
	\includegraphics[width=0.9\linewidth]{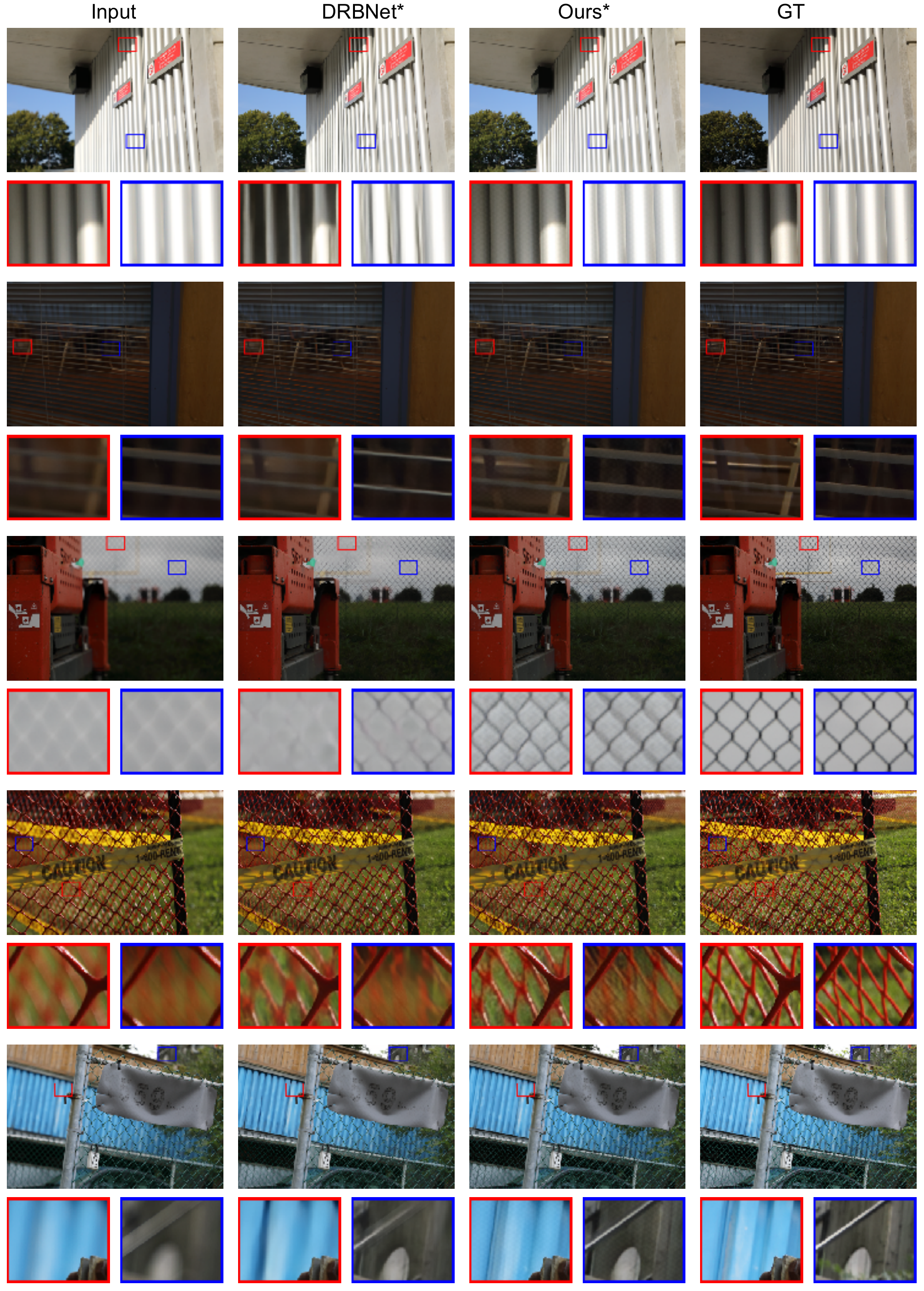}
	\caption{Qualitative comparison between DRBNet \cite{ruan2022learning} and our method adopting the two-stage training strategy as proposed in \cite{ruan2022learning} when evaluated on the DPDD dataset.}
	\label{fig:dpdd_two_stage_0}
\end{figure*}

\begin{figure*}[htb] 
	\centering
	\includegraphics[width=0.9\linewidth]{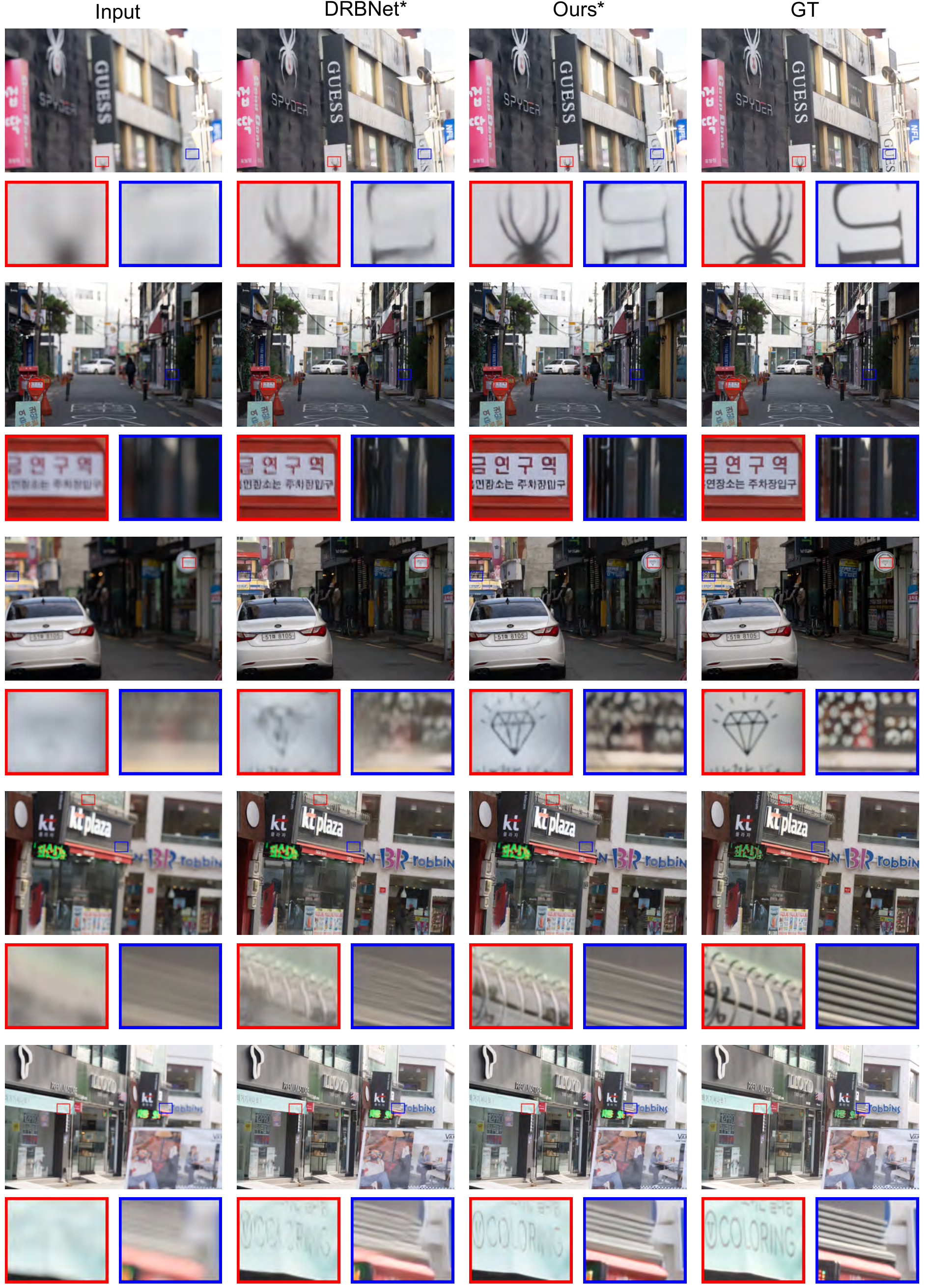}
	\caption{Qualitative comparison between DRBNet \cite{ruan2022learning} and our method adopting the two-stage training strategy as proposed in \cite{ruan2022learning} when evaluated on RealDOF dataset.}
	\label{fig:realdof_two_stage_0}
\end{figure*}

\begin{figure*}[h!] 
	\centering
	\includegraphics[width=0.7\linewidth]{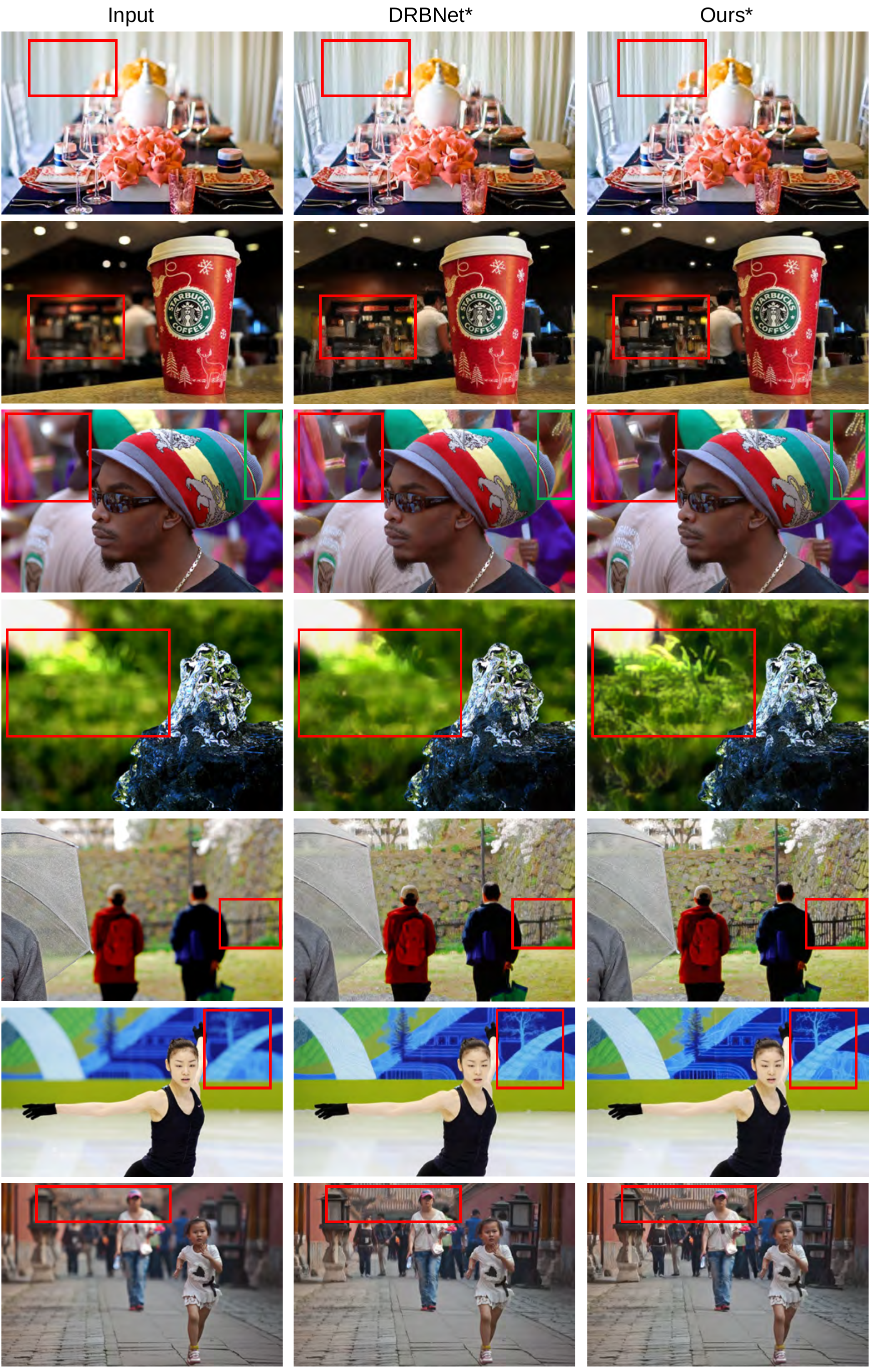}
	\caption{Qualitative comparison between DRBNet \cite{ruan2022learning} and our method evaluated on CUHK dataset when adopting the training strategy proposed in \cite{ruan2022learning}. Note no all-in-focus ground truth in CUHK dataset. }
	\label{fig:0_cuhk}
\end{figure*}

\begin{figure*}[htb] 
	\centering
	\includegraphics[width=0.7\linewidth]{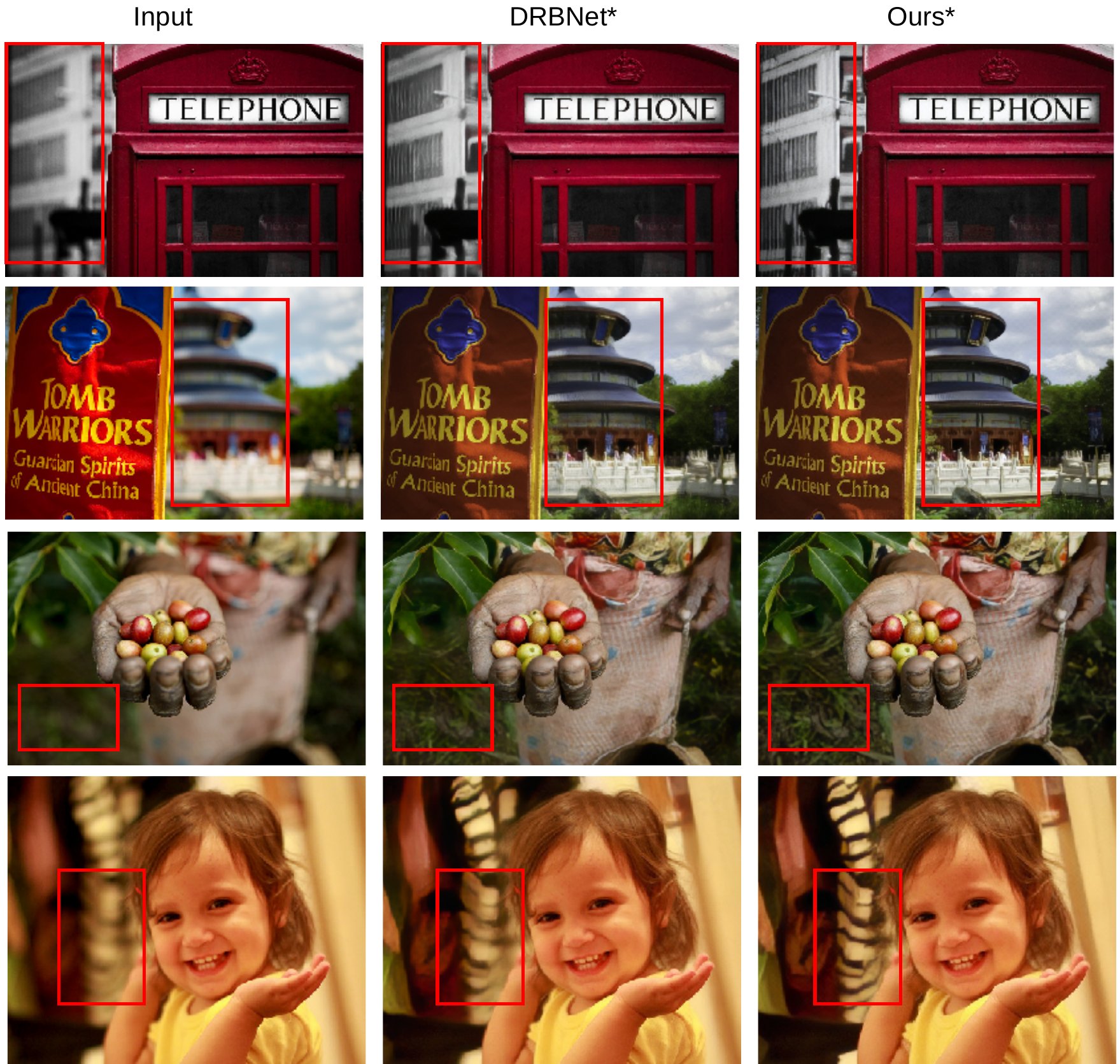}
	\caption{Qualitative comparison between DRBNet \cite{ruan2022learning} and our method evaluated on CUHK dataset when adopting the training strategy proposed in \cite{ruan2022learning}. Note no all-in-focus ground truth in the CUHK dataset.}
	\label{fig:1_cuhk}
	\vspace{-0.2cm}
\end{figure*}

\begin{figure*}[htb] 
	\centering
	\includegraphics[width=0.8\linewidth]{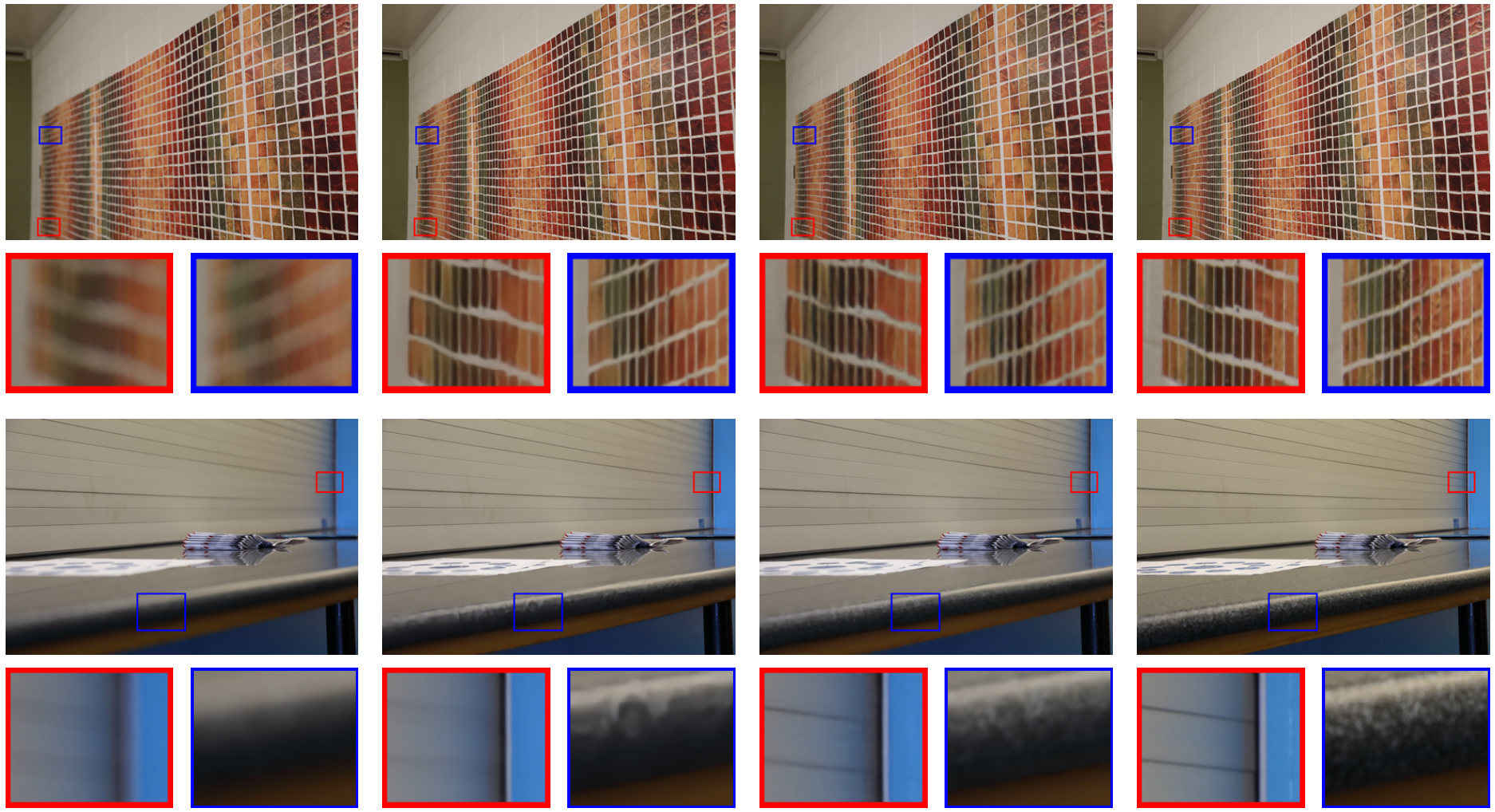}
	\caption{Dual-pixel defocus deblurring: Qualitative comparison between Restormer \cite{zamir2022restormer} and our method evaluated on DPDD dataset.}
	\label{fig:0_dual_dpdd}
\end{figure*}

\end{document}